\documentclass{article}

\usepackage{microtype}
\usepackage{graphicx}
\usepackage{subfigure}
\usepackage{booktabs} 

\usepackage{hyperref}

\usepackage[table,xcdraw]{xcolor}
\usepackage[accepted]{icml2025}

\usepackage{amsmath}
\usepackage{amssymb}
\usepackage{mathtools}
\usepackage{amsthm}
\usepackage{chngcntr} 
\usepackage{appendix} 
\usepackage[capitalize,noabbrev]{cleveref}

\theoremstyle{plain}
\newtheorem{theorem}{Theorem}[section]
\newtheorem{proposition}[theorem]{Proposition}

\newtheorem{corollary}[theorem]{Corollary}
\theoremstyle{definition}
\newtheorem{definition}[theorem]{Definition}
\newtheorem{assumption}[theorem]{Assumption}
\theoremstyle{remark}

\usepackage{listings}
\usepackage{algorithm}
\usepackage{algorithmic}
\usepackage{caption}
\usepackage{graphicx,subfigure}
\usepackage{xcolor}
\usepackage{wrapfig}
\renewcommand\algorithmiccomment[1]{\hfill $\triangleright$ #1}

\usepackage[textsize=tiny]{todonotes}

\icmltitlerunning{Revisiting Bisimulation Metric for Robust Representations in Reinforcement Learning}

\begin{document}

\twocolumn[
\icmltitle{Revisiting Bisimulation Metric for Robust Representations in Reinforcement Learning}



\icmlsetsymbol{equal}{*}

\begin{icmlauthorlist}
\icmlauthor{Leiji Zhang}{equal,bit}
\icmlauthor{Zeyu Wang}{equal,bit}
\icmlauthor{Xin Li}{bit}
\icmlauthor{Yao-Hui Li}{bit}
\end{icmlauthorlist}

\icmlaffiliation{bit}{Beijing Institute of Technology, Beijing, China}

\icmlcorrespondingauthor{Xin Li}{xinli@bit.edu.cn}

\icmlkeywords{Machine Learning, ICML}

\vskip 0.3in
]



\printAffiliationsAndNotice{\icmlEqualContribution} 

\begin{abstract}
Bisimulation metric has long been regarded as an effective control-related representation learning technique in various reinforcement learning tasks. However, in this paper, we identify two main issues with the conventional bisimulation metric: 1)
an inability to represent certain distinctive scenarios, and 2) a reliance on predefined weights for differences in rewards and subsequent states during recursive updates. 
We find that the first issue arises from an imprecise definition of the reward gap, whereas the second issue stems from overlooking the varying importance of reward difference and next-state distinctions across different training stages and task settings.
To address these issues, by introducing a measure for state-action pairs, we propose a revised bisimulation metric that features a more precise definition of reward gap and novel update operators with adaptive coefficient. We also offer theoretical guarantees of convergence for our proposed metric and its improved representation distinctiveness.
In addition to our rigorous theoretical analysis, we conduct extensive experiments on two representative benchmarks, DeepMind Control and Meta-World, demonstrating the effectiveness of our approach. The code is publicly available at: \url{https://github.com/zpwdev/RevBis}.
\end{abstract}


\addtocontents{toc}{\protect\setcounter{tocdepth}{0}}
\section{Introduction}
\label{introduction}

The application of Visual Reinforcement Learning (VRL) in real-world scenarios, such as autonomous driving, robot control, and video games, is growing rapidly~\cite{ebert2018visual,hafner2020mastering,kiran2021deep}. Yet, the issue of low sampling efficiency poses a significant barrier to its further implementation. As a result, representation learning from high-dimensional pixel inputs has emerged as a vital component in VRL. It allows agents to derive compact and informative features from raw observations, ultimately improving their learning efficiency and enabling them to select optimal actions.

Currently, a wide range of representation learning strategies serving RL are through designing auxiliary tasks, including the formulation of pixel-level reconstruction loss objectives~\cite{lee2020stochastic,yarats2021improving,yu2022mask} and the application of contrastive loss functions for heuristic augmentation pairs~\cite{laskin2020curl,stooke2021decoupling,zheng2023taco}. Nonetheless, these methods are task-agnostic, which may result in the encoder extracting task-irrelevant information from states.

To better capture task-relevant information, existing work resorts to \textit{bisimulation metric}~\cite{ferns2004metrics} to measure the behavioral similarity among states.
\textit{Bisimulation} is originally proposed to address homomorphisms in Markov Decision Processes (MDPs). It assesses equivalence between states in MDPs through a recursive process, identifying two states as equivalent if they yield the same immediate reward and share the same distribution over the next bisimilar states~\cite{givan2003equivalence}.
However, the concept of equivalence for stochastic processes presents challenges due to its strict requirement for exact agreement in transition probabilities. ~\citet{ferns2004metrics} then proposed the bisimulation metric, which quantifies the similarity between two states and can serve as a distance function to facilitate state aggregation. Such metric is often referred to as ``pessimistic'' because it focuses on worst-case discrepancies between states, which can lead to more conservative aggregations.

The $\pi$-bisimulation metric~\cite{DBLP:conf/aaai/Castro20} was introduced to measure the behavioral distance between states by focusing solely on the actions specified by a given policy $\pi$, rather than considering all possible actions. $\pi$-bisimulation metric enjoys strong convergence guarantees and the ability to focus on the behavior of interest in a policy-based way (See Theorem~\ref{theorem:bisimulation}), and has been used further as a representation learning objective, coupled with more efficient policy training~\cite{zhang2020learning, DBLP:conf/nips/CastroKPR21}.

\newpage

\begin{theorem}
\label{theorem:bisimulation}
\cite{DBLP:conf/aaai/Castro20}
Let $\mathbb{M}$ be the set of all measurements on $\mathcal{S}$. Define $\mathcal{F}^{\pi}: \mathbb{M} \rightarrow \mathbb{M}$ by
\begin{equation}
\label{pi-bisi-eq}
    \mathcal{F}^{\pi}(d)(s_i, s_j)=|r_{s_i}^\pi-r_{s_j}^\pi|+\gamma \mathcal{W}(d)\left(P_{s_i}^\pi, P_{s_j}^\pi\right),
\end{equation} 
where $s_i,s_j\in \mathcal{S}$, $r_{s_i}^\pi=\sum_{a_i\in\mathcal{A}}\pi(a_i|s_i)r_{s_i}^{a_i}$ , $P_{s_i}^\pi=\sum_{a_i\in\mathcal{A}}\pi(a_i|s_i)P_{s_i}^{a_i}$, $\gamma \in (0,1)$ is the discount factor, and $\mathcal{W}(d)$ is the Wasserstein distance with cost function $d$ between distributions.
Then $\mathcal{F}^{\pi}$ has a least fixed point $d_{\sim}^{\pi}$, and $d_{\sim}^{\pi}$ is a $\pi$-bisimulation metric.
\end{theorem}

\begin{figure}[t]
\vskip 0.2in
\begin{center}
     \includegraphics[width=0.47\textwidth]{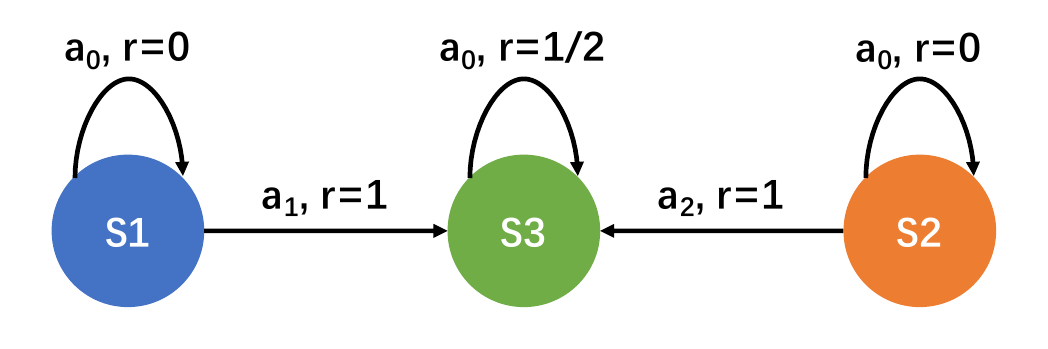}
     \caption{\textbf{Toy example.} Given a deterministic MDP $\mathcal{M} = (\mathcal{S}, \mathcal{A}, \mathcal{P}, \mathcal{R}, \gamma)$ and a policy $\pi$, where $\mathcal{S} = \{s_1, s_2, s_3\}$, $\mathcal{A} = \{a_0, a_1, a_2\}$, $\mathcal{R} = \{0, 1\}$, $\pi(a_0|s_1) = \pi(a_1|s_1) = \pi(a_0|s_2) = \pi(a_2|s_2) = \frac{1}{2}$, $\pi(a_0|s_3) = 1$, $r(s_1, a_0) = r(s_2, a_0) = 0$, $r(s_1, a_1) = r(s_2, a_2) = 1$, $r(s_3, a_0) = \frac{1}{2} $. When using $\pi$-bisimulation to calculate distance between states $s_1$ and $s_2$, $d(s_1, s_2) = 0$, which is contradictory to $\pi(\cdot|s_1) \neq \pi(\cdot|s_2)$. When using Eq.~\eqref{bisi-sa-u} and~\eqref{bisi-sa-g} to calculate distance between states $s_1$ and $s_2$, $d(s_1, s_2) = \frac{2}{(2-\gamma)(4-\gamma)}$. For detailed deductions see Appendix~\ref{explanation of toy example}.}
\vskip -0.2in
\label{toy_example}
\end{center}
\end{figure}


Despite achieving promising results, we have identified that the current bisimulation metric based methods mentioned above have two major issues:

\textbf{1) \textit{Inability to represent certain distinctive cases.}} They may fail to precisely measure the distance between states in some instances. Considering the scenarios depicted in Figure~\ref{toy_example}, if we employ $\pi$-bisimulation (as outlined in Theorem~\ref{theorem:bisimulation}) to determine the distance between states $s_1$ and $s_2$, the resulting distance $d(s_1,s_2)$ would be $0$. This suggests that $s_1$ and $s_2$ are equivalent, leading to the execution of the same policy in an abstract MDP for both states, which is contradictory to the given $\pi(\cdot|s_1) \neq \pi(\cdot|s_2)$. 
Upon analysis, we identify the issue as stemming from an imprecise definition of the reward gap in $\pi$-bisimulation Eq.~\eqref{pi-bisi-eq} which correspondingly defines the reward difference as $|\mathbb{E}_{a_i \sim \pi(\cdot|s_i)[r(s_i,a_i)]} - \mathbb{E}_{a_j \sim \pi(\cdot|s_j)[r(s_j,a_j)]}|$, where it takes the absolute difference between those two expectations i.e. the average rewards, thus overlooking the underlying shape of the reward distribution. 
In this paper, we argue that the reward difference contributing to the bisimulation metric should be set as $\mathbb{E}_{a_i \sim \pi(\cdot|s_i), a_j \sim \pi(\cdot|s_j)}|r(s_i, a_i) - r(s_j, a_j)|$, thus to capture the full distribution of rewards and how they differ across every pair of actions selected under the policy $\pi$. We will later show that such a setting allows producing more distinctive representations (See Section~\ref{sa-bisi}). 
To mitigate the computational biases arising from expectation calculations, we reformulate the bisimulation metric in a more rigorous manner, introducing a novel definition of metric and update operators measuring similarity between state-action pairs, which serves as a proxy in turn to enforce a more rigid definition of bisimulation metric among states (See more details in Section~\ref{sa-bisi} and Definition~\ref{sa-bisi-def}). We also prove the theoretical convergence guarantees of our revised definition with the accompanying proposed update operators.


\textbf{2) \textit{Stagnation of weighing components during the recursive update.}}
In the classic bisimulation formula, differences in rewards can be interpreted as short-term variations (the first term of RHS of Eq.~\eqref{pi-bisi-eq}), whereas differences in successor states represent long-term variations (the second term of RHS in Eq.~\eqref{pi-bisi-eq}). 
However, because the predefined weighing importance between these two terms remains fixed throughout training, their relative importance never adapts to the task at hand or the current stage of learning.
We posit that this importance should vary across different tasks and over the course of training within a single task. Consequently, we propose allowing the coefficient to evolve dynamically, enabling adaptive adjustments to the importance of short-term and long-term differences.
Furthermore, we theoretically and empirically demonstrate that, under certain assumptions, our newly proposed bisimulation operators together with the dynamically adapting coefficient continue to ensure convergence while further enhancing policy learning.

The revised bisimulation metric and operators can then be applied to representation learning as an auxiliary task, aiming to minimize a bisimulation loss. This facilitates the development of a more informative encoder that extracts task-relevant features for downstream RL tasks, improving both the policy and value networks. 
To evaluate our approach, we integrate proposed modifications into two popular bisimulation-based baselines, MICo~\cite{DBLP:conf/nips/CastroKPR21} and SimSR~\cite{zang2022simsr}, i.e., substituting their original operators with ours. Experimental results demonstrate the advantages of our efforts.

Our contributions can be summarized as follows:
\begin{itemize}
\item By revisiting existing bisimulation-based methods, we identify two major issues in the conventional $\pi$-bisimulation definition: an inability to capture certain scenarios and a static weighting of parameters throughout the recursive computation process, which undermines the distinctiveness and efficiency of learned representations in reinforcement learning tasks.
\item We address the identified issues by reformulating the bisimulation metric among states with a more rigorous definition, introducing a novel definition of metric measuring the similarity of state-action pairs and corresponding update operators with adaptive coefficient. Moreover, we prove the theoretical convergence guarantees of this revised metric and the discrepancy in value functions among various states is less than the distance separating them.
\item Extensive experiments on the DMC suite and Meta-World benchmark demonstrate that our reformulated bisimulation metric effectively guides representation learning in RL tasks, leading to a significant boost in policy performance.

\end{itemize}

\section{Preliminary}

\textbf{Reinforcement Learning.} We consider the underlying environment to be modeled as a Markov decision process (MDP) defined by $\mathcal{M} = (\mathcal{S}, \mathcal{A}, \mathcal{P}, \mathcal{R}, \gamma)$, where $\mathcal{S}$ is the state space, $\mathcal{A}$ the action space, $\mathcal{P}$ the transition dynamic, $\mathcal{R}$ the reward function, $\gamma$ the discount factor. The agent's objective is to achieve the highest possible expected cumulative discounted rewards by acquiring an optimal policy, which is formulated as $\mathbb{E}_{\mathcal{P}} = [\sum_{t=0}^{\infty}\gamma^{t}r_t]$. In this paper, we approximate the system's fully observed state $s$ by stacked pixel observations instead of tackling the partial-observability problem directly (a detailed explanation can be found in Appendix B of DBC~\cite{zhang2020learning}).

\section{Reformulated Bisimulation Metric with novel operators}
\label{sa-bisi}

In this section, we introduce our reformulated bisimulation metric and the involved novel bisimulation operators. We then delve into the relevant theoretical properties of these operators, highlighting their importance for robust representation learning in reinforcement learning tasks.

\begin{definition}
\label{sa-bisi-def}
Given a policy $\pi$,  $\mathcal{F}_{u}^{\pi}: \mathbb{R}^{\mathcal{S} \times \mathcal{S}} \rightarrow \mathbb{R}^{\mathcal{X} \times \mathcal{X}}$ operator and $\mathcal{F}_{g}^{\pi}: \mathbb{R}^{\mathcal{X} \times \mathcal{X}} \rightarrow \mathbb{R}^{\mathcal{S} \times \mathcal{S}}$ are defined as follows ($\mathcal{X}$ is the space of $\mathcal{S}\times\mathcal{A}$):
\begin{equation}
\label{bisi-sa-u}
\begin{split}
    (\mathcal{F}_{u}^{\pi}U)(s_i, s_j) = \mathbb{E}_{\begin{subarray}{l} a_i \sim \pi(\cdot |s_i) \\ a_j \sim \pi(\cdot |s_j) \end{subarray}} \big[ G((s_i, a_i), (s_j, a_j)) \big],
\end{split}
\end{equation}
\begin{equation}
\label{bisi-sa-g}
\begin{split}
    (\mathcal{F}_{g}^{\pi}G)((s_i, a_i), (s_j, a_j)) = |r(s_i, a_i)-r(s_j, a_j)| \\
    \quad\quad + \gamma \mathbb{E}_{\begin{subarray}{l} s_i' \sim P(\cdot|s_i, a_i) \\ s_j' \sim P(\cdot|s_j, a_j) \end{subarray}} \big[ U(s_i', s_j') \big],
\end{split}
\end{equation}
where $U$ and $G$ are metric functions used to measure the distance between two representations. 
\end{definition}
Note that in this paper,  we employ the distance functions proposed in MICo~\cite{DBLP:conf/nips/CastroKPR21} and  SimSR~\cite{zang2022simsr} to experimentally validate our proposed approach, as demonstrated in~Eq.~\eqref{mico_dis} and Eq.~\eqref{simsr_dis}, respectively. 
\begin{equation}
\label{mico_dis}
    U(x,y) = G(x,y) = \frac{\Vert x \Vert_2 + \Vert y \Vert_2}{2} + \beta \theta(x, y),
\end{equation}
\begin{equation}
\label{simsr_dis}
    U(x,y) = G(x,y) = 1 - cosine(x,y) = 1 - \frac{x^T \cdot y}{\Vert x \Vert \cdot \Vert y \Vert},
\end{equation}
where $x\in \mathbb{R}^n$ and $y\in \mathbb{R}^n$ are $n$-dimensional vectors, $\theta$ denotes the angle formed between vectors $x$ and $y$, $\beta$ is a scalar parameter. 

The above operators enable a connection between metrics derived from states and metrics derived from state-action pairs,  offering a more refined and suitable approach compared to the conventional $\pi$-bisimulation definition, as illustrated in Section~\ref{introduction}.

\textbf{\textit{Why is our proposed new definition of bisimulation metric more appropriate for guide representation learning?}} 
Recall that a key difference between our proposed metric and the conventional one lies in how we define the reward distances contributing to the metric. Specifically, we replace the absolute difference of expectations with the expectation of the absolute differences. Formally, we have:
\begin{equation}
\begin{aligned}
    \triangle_2^r =& \big|\mathbb{E}_{a_i \sim \pi(\cdot|s_i)}\big[r(s_i, a_i)\big] - \mathbb{E}_{a_j \sim \pi(\cdot|s_j)}\big[r(s_j, a_j)\big]\big| \\ 
    =& \big| \sum_{a_i \in \mathcal{A}}\pi(a_i|s_i)r(s_i, a_i) - \sum_{a_j \in \mathcal{A}}\pi(a_j|s_j)r(s_j, a_j) \big| \\
    \leq& \sum_{a_i, a_j \in \mathcal{A}} \pi(a_i|s_i)\pi(a_j|s_j) \big| r(s_i, a_i) - r(s_j, a_j) \big| \\
    =& \mathbb{E}_{\begin{subarray}{l} a_i \sim \pi(\cdot |s_i) \\ a_j \sim \pi(\cdot |s_j) \end{subarray}} \big[| r(s_i, a_i) - r(s_j, a_j) |\big] = \triangle_1^r.
\end{aligned}
\end{equation}
Aside from the arguments presented in Section~\ref{introduction}, such design enables our revised metric to differentiate cases that the conventional $\pi$-bisimulation could not. 
As shown, the conventional distance, $\triangle_2^r$ , is bounded by our proposed distance, $\triangle_1^r$; therefore $\triangle_1^r$ can be strictly larger and consequently yield a more discriminative measurement of how states differ. This heightened sensitivity means that if two states consistently lead to different reward patterns (even in subtle ways across various actions), their distance in the representation space will be more pronounced. 



The following propositions and corollaries demonstrate that the revised bisimulation metric and its operators offer theoretical guarantees in terms of the existence, uniqueness and convergence of their fixed points. Specifically, by showing that each operator is a contraction mapping over a complete metric space, we leverage Banach’s fixed-point theorem to establish the existence and uniqueness of a limiting solution.

\begin{proposition}
\label{bisi_u}
The operator $\mathcal{F}_{u}^{\pi}U$ is a contraction mapping on $\mathbb{R}^{\mathcal{S}\times\mathcal{S}}$ with respect to $L^\infty$ norm.
\begin{proof}
See Appendix \ref{proof_proposition_4.2}. 
\end{proof}
\end{proposition}

\begin{proposition}
\label{bisi_g}
The operator $\mathcal{F}_{g}^{\pi}G$ is a contraction mapping on $\mathbb{R}^{\mathcal{X}\times\mathcal{X}}$ with respect to $L^\infty$ norm.
\begin{proof}
See Appendix \ref{proof_proposition_4.3}. 
\end{proof}
\end{proposition}

The corollary that now directly arises is a consequence of Banach's fixed-point theorem, coupled with the completeness of the product space $\mathbb{R}^{\mathcal{S}\times\mathcal{S}}$ and $\mathbb{R}^{\mathcal{X}\times\mathcal{X}}$ under the $L^\infty$ norm.

\begin{corollary}

\end{corollary}

\begin{corollary}
The operator $\mathcal{F}_{g}^{\pi}$ has a unique fixed point $G^{\pi} \in \mathbb{R}^{\mathcal{X}\times\mathcal{X}}$, and any initial function $G \in \mathbb{R}^{\mathcal{X}\times\mathcal{X}}$ will converge to $G^{\pi}$ under repeated application of $\mathcal{F}_{g}^{\pi}$. 
\end{corollary}

\begin{proposition}
\label{bisi_u_and_v_bound}
For any policy $\pi$ and states $s_i, s_j \in \mathcal{S}$, we have the following guarantee:
\begin{equation}
    |V^{\pi}(s_i) - V^{\pi}(s_j)| \leq U^{\pi}(s_i, s_j).
\end{equation}
\begin{proof}
See Appendix \ref{proof_proposition_4.6}. 
\end{proof}
\end{proposition}

\begin{proposition}
\label{bisi_g_and_q_bound}
For any policy $\pi$, states $s_i, s_j \in \mathcal{S}$ and actions $a_i, a_j \in \mathcal{A}$, we have the following guarantee:
\begin{equation}
    |Q^{\pi}(s_i, a_i) - Q^{\pi}(s_j, a_j)| \leq G^{\pi}((s_i, a_i), (s_j, a_j)).
\end{equation}
\begin{proof}
See Appendix \ref{proof_proposition_4.7}. 
\end{proof}  
\end{proposition}

\section{Guided Representation Learning}
In this section, we introduce how we utilize the revised bisimulation-metric and its operators to facilitate representation learning. 


Our proposed bisimulation-based loss function is given as:
\begin{equation}
\label{loss-sac}
\begin{aligned}
     \mathcal{L}(\Theta) =& \mathbb{E}_{(s_i, a_i, r_i, s_i'), (s_j, a_j, r_j, s_j') \sim \mathcal{D}} \big[ (\mathcal{F}_{g}^{\pi}G - G)^2 \big],
\end{aligned}
\end{equation}
\begin{equation}
\label{c-f}
    \mathcal{F}_{g}^{\pi}G = (1-c) \cdot |r(s_i, a_i) - r(s_j, a_j)| + c \cdot U(\phi(s_i'), \phi(s_j')),
\end{equation}
\begin{equation}
\label{phi-psi}
    G = G(\psi(\phi(s_i), a_i), \psi(\phi(s_j), a_j)),
\end{equation}
where $\mathcal{D}$ is the replay buffer, $U$ and $G$ are specific distance functions used to measure the distance between two representations, $\Theta=\{c,\phi,\psi\}$ is a collection of networks or coefficient that need to be updated,
$c \in (0, 1)$ is a dynamically adjustable weighting coefficient,
$\phi: \mathcal{S} \rightarrow \mathcal{Z}_S$, serves as the state encoder, transforming high-dimensional observations into a compact, low-dimensional representation, $\psi: \mathcal{Z}_S \times \mathcal{A} \rightarrow \mathcal{Z}_{SA}$, termed as the state-action pair encoder. 

When training the loss function for the state encoder $\phi$ and the state-action pair encoder $\psi$, we freeze (i.e., stop gradients for) both the weight coefficient $c$ and the state encoder $\phi$, following Eq.~\eqref{c-f}. Conversely, when optimizing the loss function associated with 
$c$, the gradients for $\phi$ and $\psi$ are halted as per Eq.~\eqref{phi-psi}. See Algorithm~\ref{alg} for a detailed procedure.

Note that recognizing the relative importance of discrepancies in rewards and discrepancies in next states varies across different tasks and throughout the training stages of a single task, we introduce a dynamically adaptable weighting coefficient to balance these two differences, seen in Eq.~\eqref{c-f}. 

\begin{figure*}
\centering
\includegraphics[width=0.76\linewidth]{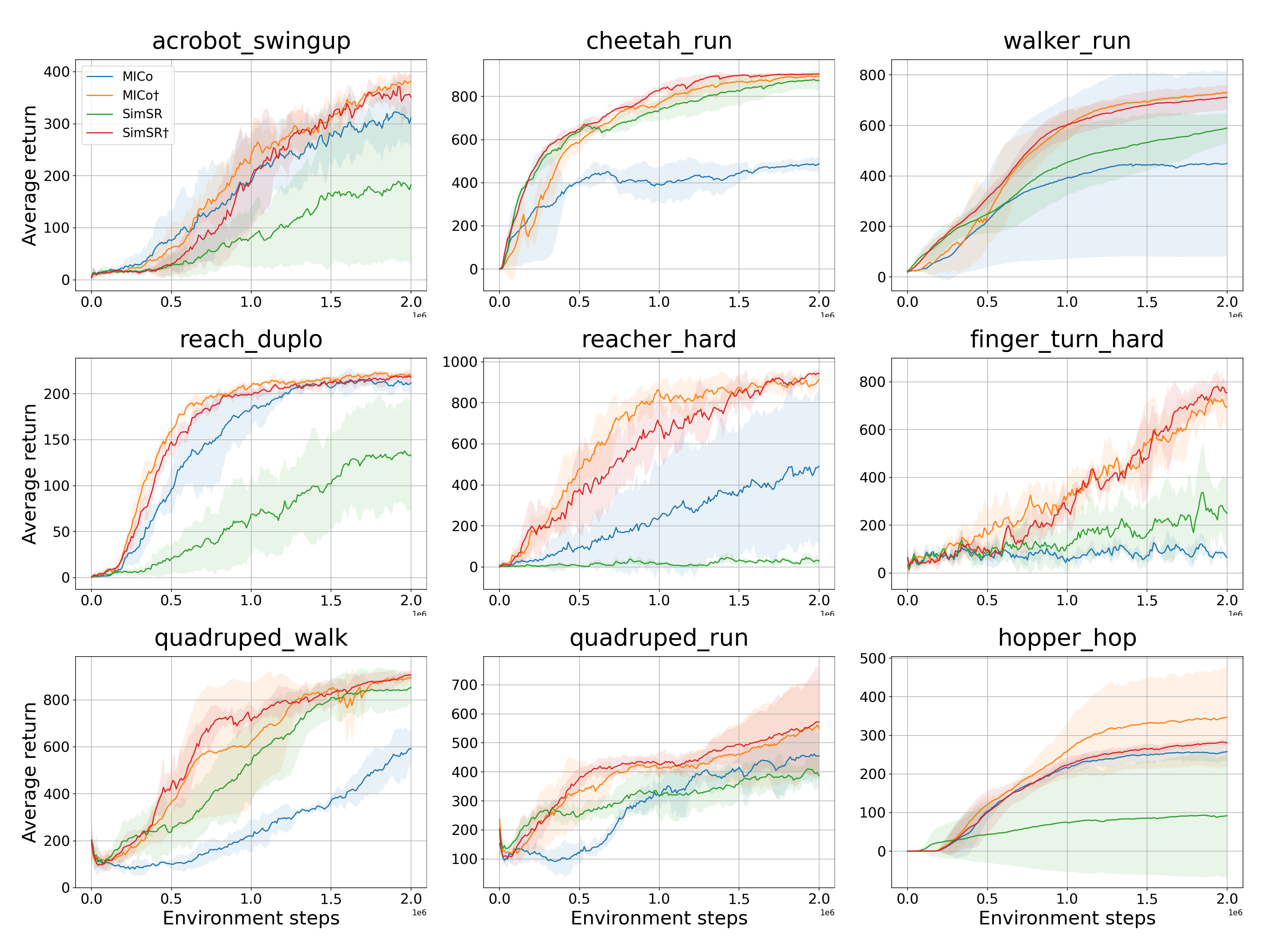}
\caption{\textbf{(Deepmind Control Suite)} Performance comparison on 9 visual-based DMC tasks over 3 seeds with one standard error shaded in the default setting. For each seed, the average return is computed every 10,000 training steps, averaging over 10 episodes. The horizontal axis indicates the number of environment steps. The vertical axis indicates the average return.}
\label{dmc_all}
\end{figure*}

Subsequently, we demonstrate that, under a suitable assumption, while $c$ evolves adaptively throughout the training process, the incorporation of this dynamic weight coefficient into the state-action bisimulation operator nevertheless guarantees theoretical convergence.

\begin{assumption}
\label{assumption_c}
    At the $n$-th time step, $c_n \in (0,1)$ represents the value of $c$. The sequence $\{c_n\}$ is either convergent or satisfies the condition $|c_n-c_{n-1}| \leq \epsilon$, where $\epsilon$ is a sufficiently small value.
\end{assumption}

\begin{proposition}
\label{c_convergence}
    Under the assumption~\ref{assumption_c}, the operator $\mathcal{F}_{g}^{\pi}G$ in Eq.~\eqref{c-f} is a contraction mapping on $\mathbb{R}^{\mathcal{X}\times\mathcal{X}}$ with respect to $L^\infty$ norm.
\begin{proof}
    See Appendix \ref{proof_c_convergence}. 
\end{proof}
\end{proposition}


\begin{algorithm}[ht]
    \caption{Revised Bisimulation Algorithm}
    \label{alg}
    \begin{algorithmic}[1]
    \FOR{Time $t=0$ to $\infty$}
        \STATE{Encode state: $\mathbf{z_t} = \phi(\mathbf{s_t})$}
        \STATE{Execute action: $\mathbf{a_t} \sim \pi(\mathbf{z_t})$}
        \STATE{Record data: $\mathcal{D} \leftarrow \mathcal{D} \cup \{ s_t, a_t, s_{t+1}, r(s_t, a_t) \}$}
        \STATE{Sample batch: $B_t \sim \mathcal{D}$}
        \STATE{Update value network: $\mathbb{E}_{\mathcal{B}_t} \big[\mathcal{L}(V) \big]$ \algorithmiccomment{DrQ-v2}}
        \STATE{Update policy network: $\mathbb{E}_{\mathcal{B}_t} \big[ \mathcal{L}(\pi)\big]$ \algorithmiccomment{DrQ-v2}}
        \STATE{Update state encoder network $\phi$ and state-action encoder network $\psi$: \\ \quad\quad\quad\quad\quad $\mathcal{L}(\phi, \psi) = \mathbb{E}_{\mathcal{B}_t} \big[ (\mathcal{F}_{g}^{\pi}\overline{G} - G)^2 \big],$\\ where $\mathcal{F}_{g}^{\pi}\overline{G} = (1 - c) \cdot |r(s_i, a_i) - r(s_j, a_j)| + c\cdot U(\overline{\phi}(s_i'), \overline{\phi}(s_j'))$, $\overline{\phi}$ indicates denotes a stop-gradient version of $\phi$.}
        \STATE{Update the weight coefficient $c$: \\ \quad\quad\quad\quad\quad $\mathcal{L}(c) = \mathbb{E}_{\mathcal{B}_t} \big[ (\mathcal{F}_{g}^{\pi}G - \overline{G})^2 \big],$ \\ where $\overline{G} = G(\overline{\psi}(\overline{\phi}(s_i), a_i), \overline{\psi}(\overline{\phi}(s_j), a_j))$, $\overline{\phi}$ and $\overline{\psi}$ indicatesindicate stop-gradient versions of $\phi, \psi$.}
    \ENDFOR
    \end{algorithmic}
\end{algorithm}

\section{Experiment}
In this section, we delve into the following pivotal questions:
1) Does the revised bisimulation operator outperform the original version?
2) How does our representation method compare to others in terms of performance?
3) What are the implications of the state-action pair encoder and the weight coefficient $c$ on our methodology?
4) How does the weight coefficient $c$ vary across different algorithms and tasks?

\subsection{Comparison between revised bisimulation operators and the original version}

\begin{figure*}
\centering
\includegraphics[width=0.80\linewidth]{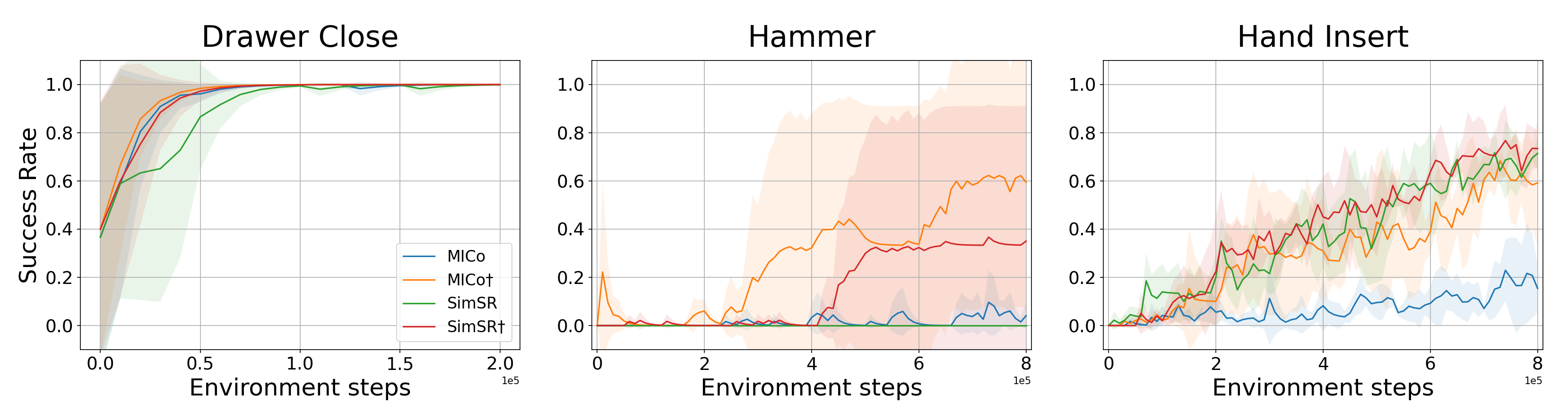}
\caption{\textbf{(Meta-World)} Performance comparison on 3 visual-based Meta-World tasks over 3 seeds with one standard error shaded in the default setting. For each seed, the success rate is computed every 10,000 training steps, averaging over 10 episodes. The horizontal axis indicates the number of environment steps. The vertical axis indicates the success rate.}
\label{metaworld}
\end{figure*}

\begin{table*}[]
\caption{\textbf{(Deepmind Control Suite)} Performance of MICo$\dagger$, SimSR$\dagger$ and other visual RL algorithms on the image-based DMControl 1M benchmark.
The results of MICo$\dagger$, SimSR$\dagger$, MICo, and SimSR are the average of 3 random seeds, while TACO, DrQ-v2, Dreamer-v3, and TDMPC are the average of 6 random seeds copied from~\citet{zheng2023taco}.
Data that is bold represents optimal results, whereas data that is underlined indicates suboptimal results.}
\label{table-dmc-all}
\resizebox{\linewidth}{!}{ 
\begin{tabular}{ccccccccc}
& \multicolumn{6}{c}{Model-free} & \multicolumn{2}{c}{Model-based} \\ \hline
\multicolumn{1}{c|}{Task (1M steps)}  &\cellcolor{gray!20} MICo$\dagger$ &\cellcolor{gray!20} SimSR$\dagger$  & MICo & SimSR  & TACO  & \multicolumn{1}{c|}{DrQ-v2} & Dreamer-v3     & TDMPC          \\ \hline\hline
\multicolumn{1}{c|}{Acrobot Swingup}  & \cellcolor{gray!20}\textbf{309 $\pm$ \textbf{59}} & \cellcolor{gray!20} \underline{248 $\pm$ 100} & 223 $\pm$ 78 & 102 $\pm$ 74 & 241 $\pm$ 21 & \multicolumn{1}{c|}{128 $\pm$ 8}  & 210 $\pm$ 12    & 224 $\pm$ 20    \\
\multicolumn{1}{c|}{Cheetah Run}  &\cellcolor{gray!20} 791 $\pm$ 26  &\cellcolor{gray!20} \textbf{848} $\pm$ \textbf{22} & 483 $\pm$ 19 & 767 $\pm$ 76 & \underline{821 $\pm$ 48} & \multicolumn{1}{c|}{691 $\pm$ 42} & 728 $\pm$ 32    & 565 $\pm$ 61    \\
\multicolumn{1}{c|}{Walker Run} &\cellcolor{gray!20} 626 $\pm$ 47 &\cellcolor{gray!20} 618 $\pm$ 64 & 403 $\pm$ 329 & 470 $\pm$ 130 & \underline{637 $\pm$ 11} & \multicolumn{1}{c|}{517 $\pm$ 43} & \textbf{765} $\pm$ \textbf{32}    & 600 $\pm$ 28    \\
\multicolumn{1}{c|}{Reach Duplo}  &\cellcolor{gray!20} \underline{217 $\pm$ 8} &\cellcolor{gray!20} 212 $\pm$ 6  & 192 $\pm$ 31 & 91 $\pm$ 64  & \textbf{234} $\pm$ \textbf{21} & \multicolumn{1}{c|}{206 $\pm$ 32} & 119 $\pm$ 30    & 117 $\pm$ 12  \\
\multicolumn{1}{c|}{Reacher Hard} &\cellcolor{gray!20}\textbf{963} $\pm$ \textbf{12} &\cellcolor{gray!20} 857 $\pm$ 93 & 307 $\pm$ 265 & 95 $\pm$ 94 & \underline{883 $\pm$ 63} & \multicolumn{1}{c|}{572 $\pm$ 51} & 499 $\pm$ 51    & 485 $\pm$ 31    \\
\multicolumn{1}{c|}{Finger Turn Hard} &\cellcolor{gray!20} 454 $\pm$ 146 &\cellcolor{gray!20} 359 $\pm$ 117 & 200 $\pm$ 100 & 243 $\pm$ 124 & \underline{632 $\pm$ 75} & \multicolumn{1}{c|}{220 $\pm$ 21} & \textbf{810} $\pm$ \textbf{58}    & 400 $\pm$ 113   \\
\multicolumn{1}{c|}{Quadruped Walk} &\cellcolor{gray!20} 652 $\pm$ 234 &\cellcolor{gray!20} \underline{786 $\pm$ 74}  & 249 $\pm$ 42 & 575 $\pm$ 125 & \textbf{793} $\pm$ \textbf{8}  & \multicolumn{1}{c|}{680 $\pm$ 52} & 353 $\pm$ 27    & 435 $\pm$ 16    \\
\multicolumn{1}{c|}{Quadruped Run} &\cellcolor{gray!20} 455 $\pm$ 14 &\cellcolor{gray!20} \underline{459 $\pm$ 18}   & 396 $\pm$ 55 & 375 $\pm$ 91 & \textbf{541} $\pm$ \textbf{38} & \multicolumn{1}{c|}{407 $\pm$ 21} & 331 $\pm$ 42    & 397 $\pm$ 37    \\
\multicolumn{1}{c|}{Hopper Hop} &\cellcolor{gray!20} \underline{274 $\pm$ 92} &\cellcolor{gray!20} 230 $\pm$ 20  & 227 $\pm$ 16 & 79 $\pm$ 136 & 261 $\pm$ 52 & \multicolumn{1}{c|}{189 $\pm$ 35} & \textbf{369 $\pm$ 21}    & 195 $\pm$ 18    \\ \hline
\end{tabular}
}
\end{table*}

\textbf{Implementation details.}
In this section of the experiment, our objective is to evaluate and compare the performance of the revised bisimulation operators against the original bisimulation operator. 
To this end, we begin with two established methods that employ the original bisimulation operator: MICo~\cite{DBLP:conf/nips/CastroKPR21} and SimSR~\cite{zang2022simsr}. Note that MICo and SimSR employ the traditional $\pi$-bisimulation operators but utilize distinct distance functions (See Eq.~\eqref{mico_dis} and~\eqref{simsr_dis}).  We then replace the operator in each method with our revised version, yielding MICo$\dagger$ and SimSR$\dagger$. 
As bisimulation-based methods can be regarded as a means of learning representations through auxiliary tasks, they allow encoders to be trained concurrently with policy and value networks. This makes them straightforward to incorporate into any model-based or model-free reinforcement learning methods as additional components.
For a fair comparison, we construct agents by integrating these four methods with the DrQ-v2~\cite{yarats2021mastering} algorithm, thereby developing practical RL methods.
We conduct experiments on two benchmarks: Deepmind Control Suite (DMC)~\cite{tassa2018deepmind} and Meta-World~\cite{yu2020meta}
. For the DMC benchmark, we choose nine challenging visual continuous control tasks, specifically: \textit{Acrobot Swingup}, \textit{Cheetah Run}, \textit{Walker Run}, \textit{Reach Duplo}, \textit{Reacher Hard}, \textit{Finger Turn Hard}, \textit{Quadruped Walk}, \textit{Quadruped Run}, and \textit{Hopper Hop}.
These tasks pose challenges due to issues such as delayed and sparse rewards, necessitating the learning of a robust representation to aid the agent in making optimal decisions.
Additionally, we select three tasks with increasing complexity (simple, medium, and hard) to analyze the algorithm’s performance across different task difficulties: \textit{Drawer Close}, \textit{Hammer}, \textit{Hand Insert}.
To facilitate the observation of experimental results, we employ the exponentially weighted moving average (EWMA) technique to smooth the data, using an attenuation coefficient of $\alpha=0.2$ in DMC and $\alpha=0.5$ in Meta-World.

\textbf{Observation 1.} 
Figure~\ref{dmc_all} compares methods employing traditional $\pi$-bisimulation operators (MICo and SimSR) with those using revised bisimulation operators (MICo$\dagger$ and SimSR$\dagger$) across nine tasks in the DMC benchmark. Meanwhile, Figure~\ref{metaworld} showcases the comparative outcomes of these methods on three tasks in the Meta-World benchmark.
Evidently, the methods incorporating the revised bisimulation operator outperform their traditional counterparts in terms of both sample efficiency and final performance. This underscores the effectiveness of the revised bisimulation operator in aiding agents in acquiring superior representations, which ultimately translates to improved performance.

\subsection{Comparison with other RL Baselines}

\textbf{Baselines.}
We compare MICo$\dagger$ and SimSR$\dagger$ against several state-of-the-art methods, with a particular focus on their respective control-related representations, including: 1) TACO~\cite{zheng2023taco}; 2) DrQ-v2~\cite{yarats2021mastering}; 3) Dreamerv3~\cite{hafner2023mastering}; 4) TDMPC~\cite{hansen2022temporal}. For clarity, we again report results for the original. 5) MICo~\cite{DBLP:conf/nips/CastroKPR21} and 6) SimSR~\cite{zang2022simsr}. TACO incorporates mutual information into representation learning, enhancing robustness by maximizing mutual information between representations of current states paired with action sequences and those of corresponding future states. DrQ-v2 is a typical model-free algorithm that leverages data augmentations to tackle challenging visual control tasks. Meanwhile, Dreamer-v3 and TDMPC are popular model-based approaches that learn world models in latent space and choose actions through model-predictive control or a learned policy.

\begin{figure*}
\centering
\includegraphics[width=0.8\linewidth]{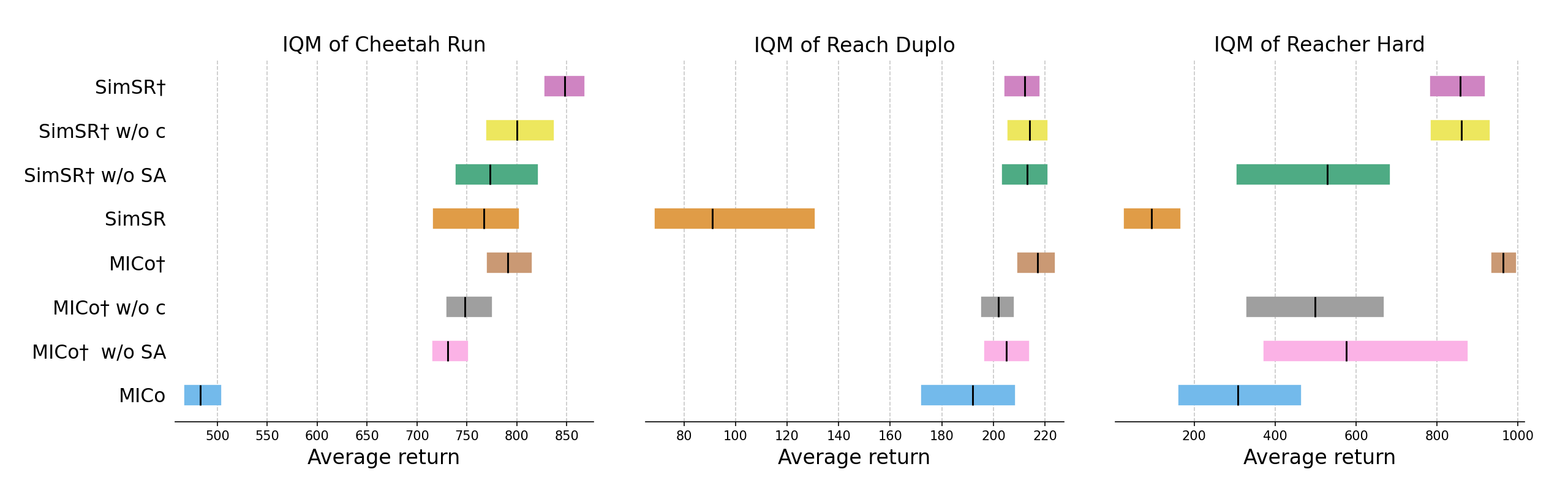}
\caption{\textbf{(Ablation Study)} Comparison results on \textit{Cheetah Run}, \textit{Reach Duplo}, and \textit{Reacher Hard} over 3 random seeds. The aggregate metric (IQM), with 95\% confidence intervals, is used to evaluate the performance. Higher IQM (i.e. inter-quartile mean) is better.
The horizontal axis indicates the average return.
The vertical axis indicates different methods. ``w/o c" denotes not introducing learnable weight coefficient $c$ and ``w/o SA" signifies the absence of the state-action pair encoder.
}
\label{ablation}
\end{figure*}

\textbf{Observation 2.}
Note that MICo$\dagger$, SimSR$\dagger$, MICo, and SimSR are all implemented by adding an auxiliary task module under the DrQ-v2 framework.
As shown in Table~\ref{table-dmc-all}, the original MICo and SimSR often lag behind DrQ-v2 (in 7 out of 9 tasks), indicating that they do not contribute positively to the agent's representation learning and ultimately hinder its final performance. In contrast, both upgraded versions MICo$\dagger$ and SimSR$\dagger$ consistently improve over DrQ-v2 in most tasks, with MICo\textdagger{} surpassing DrQ-v2 in 8 tasks and SimSR\textdagger{} outperforming it in all 9 tasks.  

When compared to TACO \citep{zheng2023taco}, both MICo\textdagger{} and SimSR\textdagger{} yield comparable or superior outcomes while relying on fewer modules. Specifically, TACO incorporates much more complex modules in their framework, i.e., contrastive learning, CURL loss, and reward prediction, and its ablation studies \citep{zheng2023taco} indicate that removing any one of them substantially leads to a significant decline in TACO's performance. By contrast, MICo\textdagger{} and SimSR\textdagger{} each rely on only a bisimulation loss and a c-loss, thus avoiding the introduction of extra parameters.

Compared to model-based methods, which require substantial resources to learn a world model, both MICo$\dagger$ and SimSR$\dagger$ consistently achieve outperformance over Dreamer-v3 in 6 tasks and TDMPC in 8 tasks. These experimental results demonstrating the efficacy of our methods.

\subsection{Ablation Study}

\textbf{Implementation details.}
To assess the impact of the state-action pair encoder and the dynamically adjustable weight coefficient $c$ on our method, we perform ablation experiments specifically targeting these components in the \textit{Cheetah Run}, \textit{Reach Duplo}, and \textit{Reacher Hard}. Figure~\ref{ablation} depicts the optimal result achieved within 1M steps, wherein MICo$\dagger$ and SimSR$\dagger$ utilize a revised bisimulation operator (detailed in Eq.~\eqref{loss-sac},~\eqref{c-f},~\eqref{phi-psi}). ``w/o c" denotes the scenario in which the revised bisimulation operator uses the RL task's specified discount factor $\gamma$, as described in Eq.~\eqref{pi-bisi-eq}, replacing the previously employed dynamically adjustable weight coefficient, while ``w/o SA" signifies removing the state-action pair encoder but still incorporating the dynamically adjustable weight coefficient $c$.

\textbf{Observation 3.}
Upon inspecting Figure~\ref{ablation}, it is evident that in all three tasks, the inclusion of a state-action pair encoder or the weight coefficient $c$ notably enhances the performance of both MICo and SimSR. Furthermore, in the majority of instances, the omission of either the state-action pair encoder or the utilization of a dynamically adjustable weight coefficient $c$ results in a decline in the performance of MICo$\dagger$ and SimSR$\dagger$. This confirms the efficacy of our proposed improvements in addressing the two primary challenges associated with traditional $\pi$-bisimulation.
\begin{figure*}[t]
\begin{center}
    \subfigure[Variation curve of coefficient $c$ under MICo$\dagger$]{
        \includegraphics[width=0.40\textwidth]{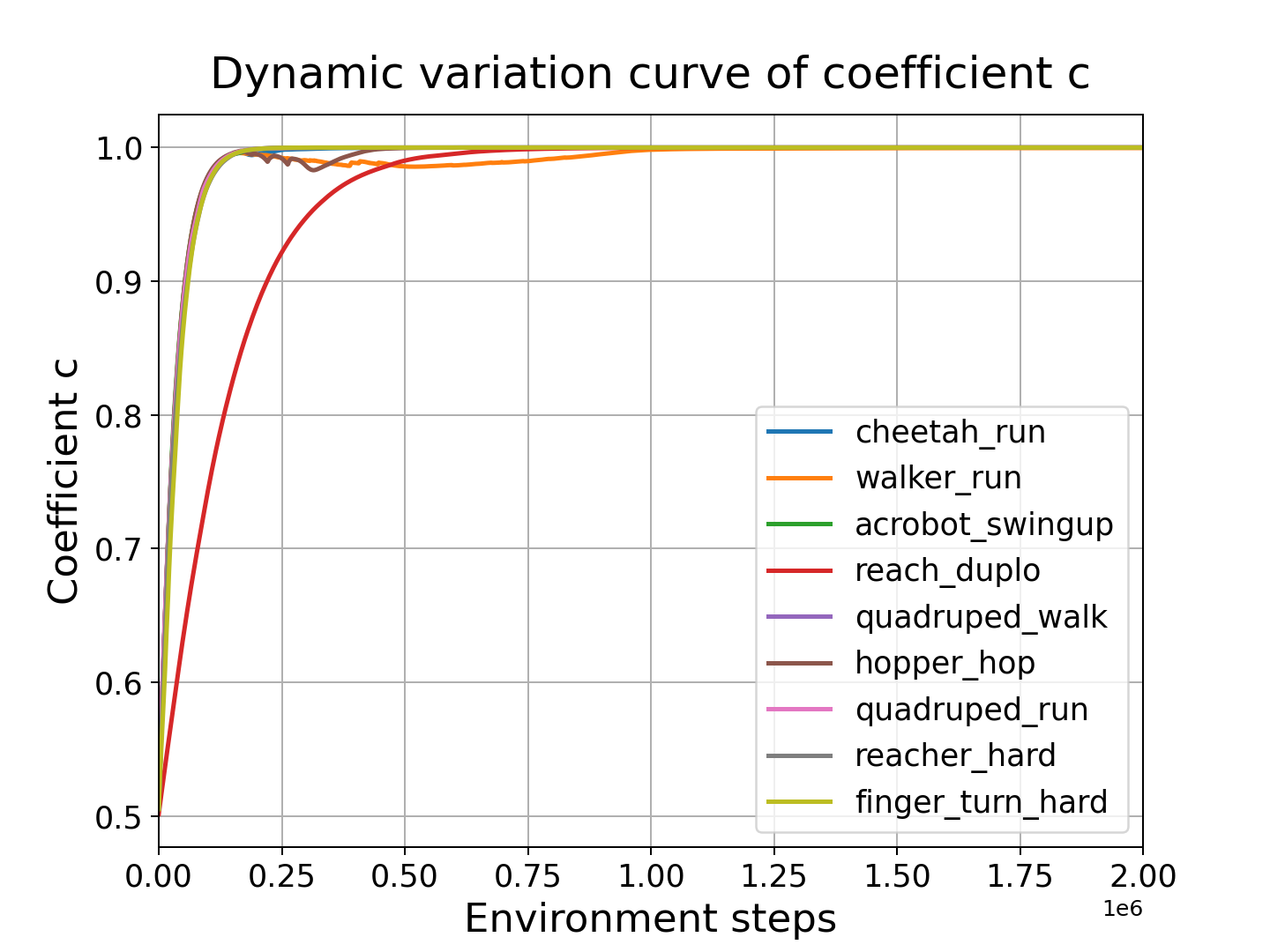}
    }
    \subfigure[Variation curve of coefficient $c$ under SimSR$\dagger$]{
        \includegraphics[width=0.40\textwidth]{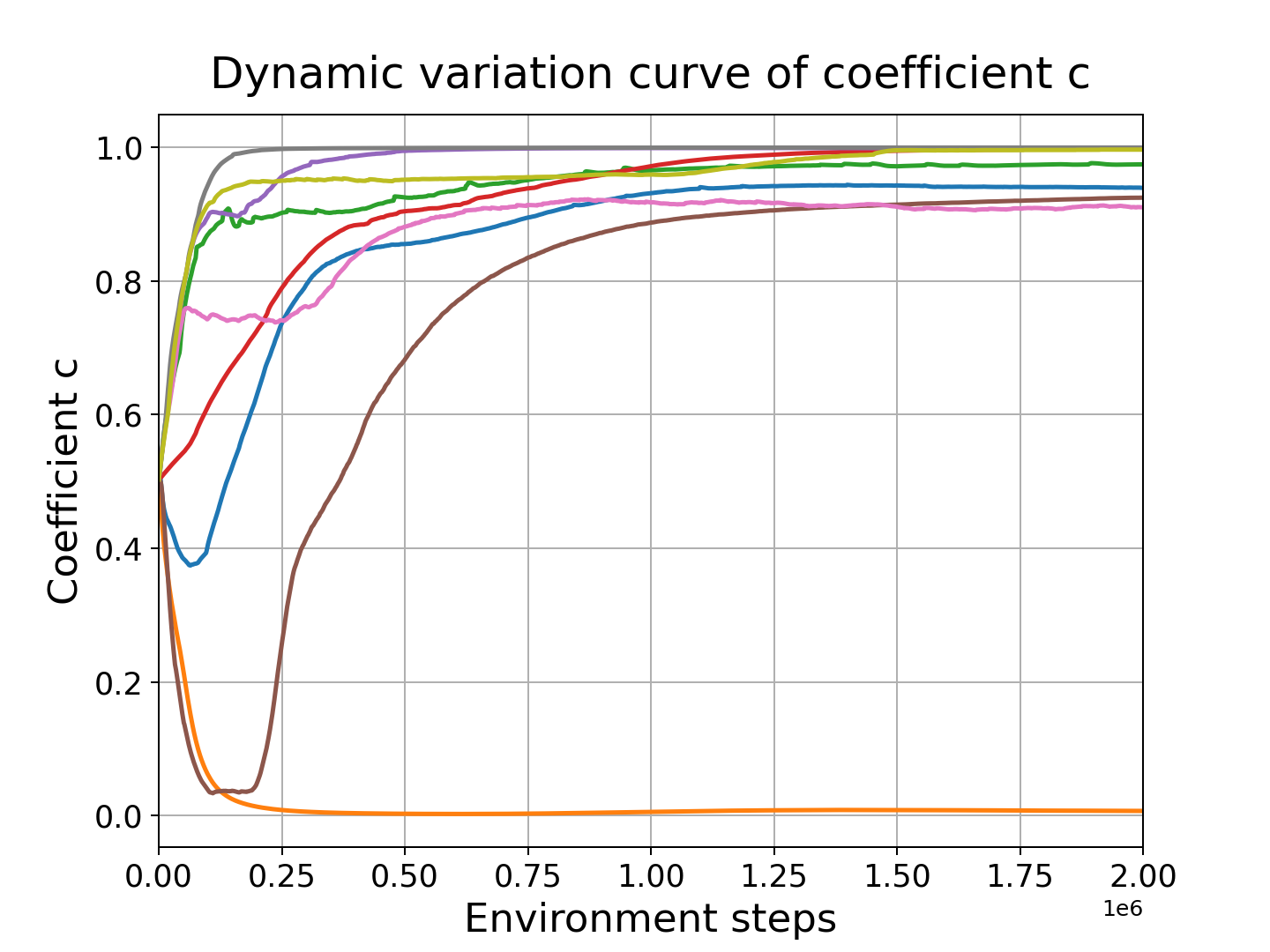}
    }
    \caption{\textbf{The dynamic variation curve of the weight coefficient $\mathbf{c}$ under MICo$\dagger$ and SimSR$\dagger$.}
    Under MICo$\dagger$, the weight coefficient $c$ displays an upward trend and eventually converges within each task. However, under SimSR$\dagger$, the weight coefficient $c$ exhibits diverse trends of change across various tasks.
    }
\label{curve_c}
\end{center}
\end{figure*}

\subsection{Evolution Behavior of the coefficient $\mathbf{c}$}

\textbf{Implementation details.}
To observe the evolution of the weight coefficient $c$, we initialize $c$ to 0.5 and subsequently record its value every 250 steps throughout the training process of each task. Subsequently, These recorded values are then plotted in a curve, as shown in Figure~\ref{curve_c}.

\textbf{Observation 4.}
In the bisimulation-based method, the difference in immediate rewards can be regarded as a short-term difference between states, whereas the difference in subsequent states signifies a long-term difference. From a training perspective, we argue that the significance of these short-term and long-term differences can shift across different training stages and tasks w.r.t. different bisimulation-based methods, thus necessitating a dynamic balancing mechanism to appropriately weigh these factors. Figure~\ref{curve_c} showcases how the weight coefficient $c$ evolves during training across multiple tasks for both the MICo$\dagger$ and SimSR$\dagger$. First, we observe clear convergence for both methods across all nine tasks. These experimental findings verify that our Assumption~\ref{assumption_c} is practical to guarantee the convergence of finding the appropriate representations under the impact of the revised operations and the dynamic weight coefficient. Second, distinct patterns emerge for each method: under MICo$\dagger$, $c$ consistently increases until convergence in all nine tasks, whereas under the SimSR$\dagger$ method, the weight coefficient $c$ exhibits three distinct trends among the nine tasks: a continual increase until convergence, a continuous decline until convergence, and an initial decrease followed by an increase until convergence. We believe these evolution behavior patterns are due to the different distance functions adopted by each method and the reward configurations of these tasks, both of which can influence how $c$ adapts throughout training.  

\section{Related Work}

\textbf{Visual Representation Learning in RL.} 
Visual reinforcement learning holds considerable importance in practical applications, including autonomous driving, robotic control, video games, etc. Nonetheless, the presence of extraneous or irrelevant details in high-dimensional observations can make it highly challenging for RL agents to learn effective representations. 
One prevalent method for learning effective representations involves optimizing both the policy and auxiliary objectives simultaneously. 
~\citet{lee2020stochastic, yarats2021improving, yu2022mask} focuses on utilizing deep auto-encoders with reconstruction loss to extract low-dimensional representations and enhance policy performance in visual reinforcement learning.
~\citet{guo2020bootstrap, schwarzer2020data, yu2021playvirtual} identify useful information from images by predicting rewards or dynamics, thereby facilitating policy learning.
~\citet{DBLP:conf/aaai/Castro20, zhang2020learning, DBLP:conf/nips/CastroKPR21, zang2022simsr} apply the bisimulation metric to aggregate states, effectively mitigating the disruption caused by extraneous information in images.
~\citet{laskin2020curl, stooke2021decoupling, zheng2023taco, yan2024enhancing} employ the contrastive loss function to minimize the distance between similar states and maximize the distance between different states in the latent space, thereby enabling the learning of an effective representation.
Additionally, several data augmentation techniques, such as overlay~\cite{zhang2017mixup}, color jitter~\cite{cubuk2019autoaugment}, random cropping~\cite{yarats2021mastering}, have demonstrated significant benefits for enhancing visual representation learning in RL.

\textbf{Bisimulation Based Methods.}
The concept of bisimulation was first introduced by~\citet{larsen1989bisimulation}, which defines that two states of a process are considered equivalent if all transitions from one state can be mirrored by transitions from the other, with the resulting states themselves being bisimilar. 
Building upon this, ~\citet{givan2003equivalence} extended bisimulation to transition systems with rewards in the context of MDP. However, the concept of equivalence for stochastic processes presents challenges due to its strict requirement for exact agreement in transition probabilities. To address this,~\citet{ferns2004metrics} proposed bisimulation metrics, which quantify the similarity between two states and can serve as a distance function to facilitate state aggregation. Unfortunately, this metric tends to be overly pessimistic by considering worst-case discrepancies between states. To overcome this limitation,~\citet{DBLP:conf/aaai/Castro20} introduced an on-policy variant of bisimulation that focuses on the behavior of interest.
Despite these advancements, bisimulation still incurs substantial computational complexity due to the need to calculate the distance between state transition distributions. To mitigate this issue, DBC~\cite{zhang2020learning} assumes that the state transition distribution conforms to a Gaussian distribution and employs the 2-Wasserstein metric to determine the $W_2$ distance between states, enabling a closed-form solution for computing the Wasserstein distance. Alternatively, MICo~\cite{DBLP:conf/nips/CastroKPR21} avoids the calculation of the Wasserstein distance between two state transition distributions altogether by utilizing an independent couple sampling strategy.
However, this design has the potential to suffer from representation collapse, as it violates the ``zero self-distance" property. 
To address this concern,~\citet{zang2022simsr} introduced SimSR, which successfully resolves the issue.

\section{Conclusion}
In this work, we revisit the bisimulation metric and reveal two key limitations: the inability to capture certain distinct scenarios and the oversight of how reward and subsequent-state differences hold varying importance during recursive updates. To address these issues, we refine prior $\pi$-bisimulation definitions with novel operators and provide a detailed theoretical analysis. Our experiments demonstrate that simply substituting the original metric with our newly proposed metric in existing bisimulation-based approaches significantly improves policy performance, even surpassing several more complex state-of-the-art models.

\newpage
\section*{Impact Statement}
This paper presents work whose goal is to advance the field of Reinforcement Learning. There are many potential societal consequences of our work, none of which we feel must be specifically highlighted here.

\bibliography{example_paper}
\bibliographystyle{icml2025}

\newpage
\appendix
\onecolumn

\section{Theoretical Analyses}
\subsection{Proof of Proposition~\ref{bisi_u}}
\label{proof_proposition_4.2}

\begin{proposition}
The operator $\mathcal{F}_{u}^{\pi}U$ is a contraction mapping on $\mathbb{R}^{\mathcal{S}\times\mathcal{S}}$ with respect to $L^\infty$ norm.
\begin{proof}
\begin{equation}
\begin{split}
    &\left|(\mathcal{F}_{u}^{\pi}U)(s_i, s_j) - (\mathcal{F}_{u}^{\pi}U')(s_i, s_j)\right| \\
    = & \left|\mathbb{E}_{\begin{subarray}{l} a_i \sim \pi(\cdot |s_i) \\ a_j \sim \pi(\cdot |s_j) \end{subarray}} \big[ G((s_i, a_i), (s_j, a_j)) \big] -  \mathbb{E}_{\begin{subarray}{l} a_i \sim \pi(\cdot |s_i) \\ a_j \sim \pi(\cdot |s_j) \end{subarray}} \big[ G'((s_i, a_i), (s_j, a_j)) \big]\right| \\ 
    = & \left| \mathbb{E}_{\begin{subarray}{l} a_i \sim \pi(\cdot |s_i) \\ a_j \sim \pi(\cdot |s_j) \end{subarray}} \big[ G((s_i, a_i), (s_j, a_j)) - G'((s_i, a_i), (s_j, a_j)) \big] \right| \\ 
    = & \left| \mathbb{E}_{\begin{subarray}{l} a_i \sim \pi(\cdot |s_i) \\ a_j \sim \pi(\cdot |s_j) \end{subarray}} \left[ |r(s_i, a_i)-r(s_j, a_j)| + \gamma \mathbb{E}_{\begin{subarray}{l} s_i' \sim P(\cdot|s_i, a_i) \\ s_j' \sim P(\cdot|s_j, a_j) \end{subarray}} \big[ U(s_i', s_j') \big] - |r(s_i, a_i)-r(s_j, a_j)| - \gamma \mathbb{E}_{\begin{subarray}{l} s_i' \sim P(\cdot|s_i, a_i) \\ s_j' \sim P(\cdot|s_j, a_j) \end{subarray}} \big[ U'(s_i', s_j') \big] \right]\right| \\ 
    = & \left| \mathbb{E}_{\begin{subarray}{l} a_i \sim \pi(\cdot |s_i) \\ a_j \sim \pi(\cdot |s_j) \end{subarray}} \left[ \gamma \mathbb{E}_{\begin{subarray}{l} s_i' \sim P(\cdot|s_i, a_i) \\ s_j' \sim P(\cdot|s_j, a_j) \end{subarray}} \big[ U(s_i', s_j') \big] - \gamma \mathbb{E}_{\begin{subarray}{l} s_i' \sim P(\cdot|s_i, a_i) \\ s_j' \sim P(\cdot|s_j, a_j) \end{subarray}} \big[ U'(s_i', s_j')\big] \right] \right| \\ 
    = & \gamma \left| \mathbb{E}_{\begin{subarray}{l} a_i \sim \pi(\cdot |s_i) \\ a_j \sim \pi(\cdot |s_j) \end{subarray}} \left[ \mathbb{E}_{\begin{subarray}{l} s_i' \sim P(\cdot|s_i, a_i) \\ s_j' \sim P(\cdot|s_j, a_j) \end{subarray}} \big[ (U - U')(s_i', s_j') \big] \right] \right| \\
    \leq & \gamma\left\Vert U - U' \right\Vert_{\infty}.
\end{split}
\end{equation}
\end{proof}
\end{proposition}

\subsection{Proof of Proposition~\ref{bisi_g}}
\label{proof_proposition_4.3}

\begin{proposition}
The operator $\mathcal{F}_{g}^{\pi}G$ is a contraction mapping on $\mathbb{R}^{\mathcal{X}\times\mathcal{X}}$ with respect to $L^\infty$ norm.
\begin{proof}
\begin{equation}
\begin{split}
    &\left|(\mathcal{F}_{g}^{\pi}G)((s_i, a_i), (s_j, a_j)) - (\mathcal{F}_{g}^{\pi}G')((s_i, a_i), (s_j, a_j))\right| \\ 
    = &\left| |r(s_i, a_i)-r(s_j, a_j)| + \gamma \mathbb{E}_{\begin{subarray}{l} s_i' \sim P(\cdot|s_i, a_i) \\ s_j' \sim P(\cdot|s_j, a_j) \end{subarray}} \big[ U(s_i', s_j') \big] - |r(s_i, a_i)-r(s_j, a_j)| - \gamma \mathbb{E}_{\begin{subarray}{l} s_i' \sim P(\cdot|s_i, a_i) \\ s_j' \sim P(\cdot|s_j, a_j) \end{subarray}} \big[ U'(s_i', s_j') \big]\right| \\
    = &\left| \gamma \mathbb{E}_{\begin{subarray}{l} s_i' \sim P(\cdot|s_i, a_i) \\ s_j' \sim P(\cdot|s_j, a_j) \end{subarray}} \big[ U(s_i', s_j') \big] - \gamma \mathbb{E}_{\begin{subarray}{l} s_i' \sim P(\cdot|s_i, a_i) \\ s_j' \sim P(\cdot|s_j, a_j) \end{subarray}} \big[ U'(s_i', s_j') \big]\right| \\
    = &\gamma \left|\mathbb{E}_{\begin{subarray}{l} s_i' \sim P(\cdot|s_i, a_i) \\ s_j' \sim P(\cdot|s_j, a_j) \end{subarray}} \big[U(s_i', s_j') - U'(s_i', s_j') \big]\right| \\
    = &\gamma \left| \mathbb{E}_{\begin{subarray}{l} s_i' \sim P(\cdot|s_i, a_i) \\ s_j' \sim P(\cdot|s_j, a_j) \end{subarray}} \left[ \mathbb{E}_{\begin{subarray}{l} a_i' \sim \pi(\cdot |s_i') \\ a_j' \sim \pi(\cdot |s_j') \end{subarray}} \big[ G((s_i', a_i'), (s_j', a_j')) \big] - \mathbb{E}_{\begin{subarray}{l} a_i' \sim \pi(\cdot |s_i') \\ a_j' \sim \pi(\cdot |s_j') \end{subarray}} \big[ G'((s_i', a_i'), (s_j', a_j'))\big] \right] \right| \\
    = &\gamma \left| \mathbb{E}_{\begin{subarray}{l} s_i' \sim P(\cdot|s_i, a_i) \\ s_j' \sim P(\cdot|s_j, a_j) \end{subarray}} \left[ \mathbb{E}_{\begin{subarray}{l} a_i' \sim \pi(\cdot |s_i') \\ a_j' \sim \pi(\cdot |s_j') \end{subarray}} \big[ (G - G')((s_i', a_i'), (s_j', a_j')) \big] \right] \right| \\
    \leq& \gamma \left\Vert G - G' \right\Vert_{\infty}.
\end{split}
\end{equation}
\end{proof}
\end{proposition}

\subsection{Proof of Proposition~\ref{bisi_u_and_v_bound}}
\label{proof_proposition_4.6}

\begin{proposition}
For any policy $\pi$ and states $s_i, s_j \in \mathcal{S}$, we have the following guarantee:
\begin{equation}
    |V^{\pi}(s_i) - V^{\pi}(s_j)| \leq U^{\pi}(s_i, s_j).
\end{equation}
\begin{proof}
By using a coinductive argument as outlined in~\cite{kozen2006coinductive}, we demonstrate that if a certain condition $|V^{\pi}(s_i) - V^{\pi}(s_j)| \leq U^{\pi}(s_i, s_j)$ holds for all $s_i, s_j \in \mathcal{S}$, given a symmetric function $U \in \mathbb{R}^{\mathcal{S}\times\mathcal{S}}$ in its two arguments, then the condition $|V^{\pi}(s_i) - V^{\pi}(s_j)| \leq (\mathcal{F}_{u}^{\pi}U)(s_i, s_j)$ also holds for all $s_i, s_j \in \mathcal{S}$. Given that the hypothesis is satisfied for the constant function $U(s_i, s_j) = 2 \max_{s, a} |r(s, a)| / (1 - \gamma)$, and $\mathcal{F}_{u}^{\pi}$ contracts around $U^{\pi}$, the conclusion can be derived. Therefore, assuming the hypothesis is true, we have:
\begin{equation}
\begin{aligned}
    |V^{\pi}(s_i) - V^{\pi}(s_j)| &= \left| \sum_{a_i \in \mathcal{A}} \pi(a_i|s_i)\left(r(s_i, a_i) + \gamma \sum_{s_i' \in \mathcal{S}} P(s_i'|s_i, a_i)V^{\pi}(s_i')\right) \right. \\
    & \left. \quad\quad - \sum_{a_j \in \mathcal{A}} \pi(a_j|s_j)\left(r(s_j, a_j) + \gamma \sum_{s_j' \in \mathcal{S}} P(s_j'|s_j, a_j)V^{\pi}(s_j')\right) \right| \\
    & \leq \sum_{a_i, a_j \in \mathcal{A}}\pi(a_i|s_i)\pi(a_j|s_j)\\
    & \quad\quad \left| r(s_i, a_i) - r(s_j, a_j) + \gamma \sum_{s_i' \in \mathcal{S}} P(s_i'|s_i, a_i)V^{\pi}(s_i') - \gamma \sum_{s_j' \in \mathcal{S}} P(s_j'|s_j, a_j)V^{\pi}(s_j') \right| \\
    & \leq \sum_{a_i, a_j \in \mathcal{A}}\pi(a_i|s_i)\pi(a_j|s_j)\\
    & \quad\quad \left(\left| r(s_i, a_i) - r(s_j, a_j) \right| + \left| \gamma \sum_{s_i' \in \mathcal{S}} P(s_i'|s_i, a_i)V^{\pi}(s_i') - \gamma \sum_{s_j' \in \mathcal{S}} P(s_j'|s_j, a_j)V^{\pi}(s_j') \right|\right) \\
    & \leq \sum_{a_i, a_j \in \mathcal{A}}\pi(a_i|s_i)\pi(a_j|s_j) \\
    & \quad\quad \left( \left|r(s_i, a_i) - r(s_j, a_j)\right| + \gamma \sum_{s_i', s_j' \in \mathcal{S}} P(s_i'|s_i, a_i) P(s_j'|s_j, a_j) \left| V^{\pi}(s_i') - V^{\pi}(s_j') \right|\right) \\
    & \leq \sum_{a_i, a_j \in \mathcal{A}}\pi(a_i|s_i)\pi(a_j|s_j) \\
    & \quad\quad \left( \left|r(s_i, a_i) - r(s_j, a_j)\right| + \gamma \sum_{s_i', s_j' \in \mathcal{S}} P(s_i'|s_i, a_i) P(s_j'|s_j, a_j) U^{\pi}(s_i', s_j') \right) \\
    &= \sum_{a_i, a_j \in \mathcal{A}}\pi(a_i|s_i)\pi(a_j|s_j) G^{\pi}((s_i, a_i), (s_j, a_j)) \\
    &= (\mathcal{F}_{u}^{\pi}U)(s_i, s_j).
\end{aligned}
\end{equation}
Due to the property of symmetry, $|V^{\pi}(s_j) - V^{\pi}(s_i)| \leq U^{\pi}(s_i, s_j)$ also holds, which fulfills the requirement.
\end{proof}
\end{proposition}

\subsection{Proof of Proposition~\ref{bisi_g_and_q_bound}}
\label{proof_proposition_4.7}

\begin{proposition}
By using a coinductive argument as outlined in~\cite{kozen2006coinductive}, we demonstrate that if a certain condition $|Q^{\pi}(s_i, a_i) - Q^{\pi}(s_j, a_j)| \leq G^{\pi}((s_i, a_i), (s_j, a_j))$ holds for all $(s_i, a_i), (s_j, a_j) \in \mathcal{X}$, given a symmetric function $G \in \mathbb{R}^{\mathcal{X}\times\mathcal{X}}$ in its two arguments, then the condition $|Q^{\pi}(s_i, a_i) - Q^{\pi}(s_j, a_j)| \leq (\mathcal{F}_{g}^{\pi}G)((s_i, a_i), (s_j, a_j))$ also holds for all $(s_i, a_i), (s_j, a_j) \in \mathcal{X}$. Given that the hypothesis is satisfied for the constant function $G((s_i, a_i), (s_j, a_j)) = 2 \max_{s, a} |r(s, a)| / (1 - \gamma)$, and $\mathcal{F}_{g}^{\pi}$ contracts around $G^{\pi}$, the conclusion can be derived. Therefore, assuming the hypothesis is true, we have:
\begin{equation}
    |Q^{\pi}(s_i, a_i) - Q^{\pi}(s_j, a_j)| \leq G^{\pi}((s_i, a_i), (s_j, a_j)).
\end{equation}
\begin{proof}
The proof process of proposition~\ref{bisi_g_and_q_bound} is similar to that of proposition~\ref{bisi_u_and_v_bound}, we only need to prove:
\begin{equation}
\begin{aligned}
    |Q^{\pi}(s_i, a_i) - Q^{\pi}(s_j, a_j)| &= \left| r(s_i, a_i) - r(s_j, a_j) + \gamma \sum_{s_i' \in \mathcal{S}} \sum_{a_i' \in \mathcal{A}}P(s_i'|s_i, a_i) \pi(a_i'|s_i') Q^{\pi}(s_i', a_i') \right. \\ 
    & \quad\quad\quad\quad\quad\quad\quad\quad\quad\quad \left. - \gamma \sum_{s_j' \in \mathcal{S}} \sum_{a_j' \in \mathcal{A}}P(s_j'|s_j, a_j) \pi(a_j'|s_j') Q^{\pi}(s_j', a_j') \right| \\
    &\leq |r(s_i, a_i) - r(s_j, a_j)| \\
    & \quad\quad + \left| \gamma \sum_{s_i' \in \mathcal{S}} \sum_{a_i' \in \mathcal{A}}P(s_i'|s_i, a_i) \pi(a_i'|s_i') Q^{\pi}(s_i', a_i') - \gamma \sum_{s_j' \in \mathcal{S}} \sum_{a_j' \in \mathcal{A}}P(s_j'|s_j, a_j) \pi(a_j'|s_j') Q^{\pi}(s_j', a_j') \right| \\
    &\leq |r(s_i, a_i) - r(s_j, a_j)| \\
    & \quad\quad + \gamma \sum_{s_i', s_j' \in \mathcal{S}} \sum_{a_i', a_j' \in \mathcal{A}} P(s_i'|s_i, a_i) P(s_j'|s_j, a_j) \pi(a_i'|s_i') \pi(a_j'|s_j') \left|Q^{\pi}(s_i', a_i') - Q^{\pi}(s_j', a_j') \right| \\
    &\leq |r(s_i, a_i) - r(s_j, a_j)| \\
    & \quad\quad + \gamma \sum_{s_i', s_j' \in \mathcal{S}} \sum_{a_i', a_j' \in \mathcal{A}} P(s_i'|s_i, a_i) P(s_j'|s_j, a_j) \pi(a_i'|s_i') \pi(a_j'|s_j') G^{\pi}((s_i', a_i'), (s_j', a_j')) \\
    &= |r(s_i, a_i) - r(s_j, a_j)| + \gamma \sum_{s_i', s_j' \in \mathcal{S}} P(s_i'|s_i, a_i) P(s_j'|s_j, a_j) U^{\pi}(s_i', s_j') \\
    &= (\mathcal{F}_{g}^{\pi}G)((s_i, a_i), (s_j, a_j)).
\end{aligned}
\end{equation}
Due to the property of symmetry, $|Q^{\pi}(s_j, a_j) - Q^{\pi}(s_i, a_i)| \leq G^{\pi}((s_i, a_i), (s_j, a_j))$ also holds, which fulfills the requirement.
\end{proof}  
\end{proposition}

\subsection{Proof of Proposition~\ref{c_convergence}}
\label{proof_c_convergence}

\begin{assumption}
    At the $n$-th time step, $c_n \in (0,1)$ represents the value of $c$. The sequence $\{c_n\}$ is either convergent or satisfies the condition $|c_n-c_{n-1}| \leq \epsilon$, where $\epsilon$ is a sufficiently small value.
\end{assumption}

\begin{proposition}
    Under the assumption~\ref{assumption_c}, the operator $\mathcal{F}_{g}^{\pi}G$ in Eq.~\eqref{c-f} is a contraction mapping on $\mathbb{R}^{\mathcal{X}\times\mathcal{X}}$ with respect to $L^\infty$ norm.
\begin{proof}
Consider the sequence $\{G_n\}$ defined by:
\begin{equation}
    G_{n+1} = \mathcal{F}_{g}^{\pi}G_n^{c_n}.
\end{equation}

We need to show that $\{G_n\}$ converges.
\begin{equation}
\begin{aligned}
    |G_{n+1} - G_n| &= |\mathcal{F}_{g}^{\pi}G_n^{c_n} - \mathcal{F}_{g}^{\pi}G_{n-1}^{c_{n-1}}| \\
    &\leq |\mathcal{F}_{g}^{\pi}G_n^{c_n} - \mathcal{F}_{g}^{\pi}G_{n}^{c_{n-1}}| + |\mathcal{F}_{g}^{\pi}G_n^{c_{n-1}} - \mathcal{F}_{g}^{\pi}G_{n-1}^{c_{n-1}}|.
\end{aligned}
\end{equation}

For the first term:
\begin{equation}
\label{21}
\begin{aligned}
    &|\mathcal{F}_{g}^{\pi}G_n^{c_n} - \mathcal{F}_{g}^{\pi}G_{n-1}^{c_{n-1}}| \\
    =&\left| (1 - c_n) |r_i - r_j| + c_n \mathbb{E}[U_n^{c_n}] - (1 - c_{n-1})|r_i - r_j| - c_{n-1} \mathbb{E}[U_n^{n-1}] \right| \\
    =&\left| (c_{n-1} - c_n)|r_i - r_j| + \mathbb{E}[c_n U_n^{c_n} - c_{n-1} U_n^{c_{n-1}}] \right| \\
    =&\left| (c_{n-1} - c_n)|r_i - r_j| + \mathbb{E} 
    \left[\mathbb{E}\left[c_n G_n^{c_n} - c_{n-1} G_n^{c_{n-1}}\right] \right] \right| \\
    =& \left| (c_{n-1} - c_n)|r_i - r_j| + \mathbb{E} 
    \left[\mathbb{E}\left[ c_n(1-c_n)|r_i' - r_j'| + c_n^2 \mathbb{E}[U_n^{c_n}] - c_{n-1}(1-c_{n-1})|r_i' - r_j'| - c_{n-1}^2 \mathbb{E}[U_{n-1}^{c_{n-1}}] \right] \right]\right| \\
    =& \left| (c_{n-1} - c_n)|r_i' - r_j'| + \mathbb{E} 
    \left[\mathbb{E}\left[ (c_{n-1} - c_n)(c_n + c_{n-1} - 1)|r_i' - r_j'| + \mathbb{E}\left[ \mathbb{E}\left[ c_n^2 G_n^{c_n} - c_{n-1}^2 G_{n-1}^{c_{n-1}} \right] \right] \right] \right]\right|, \\
\end{aligned}
\end{equation}
where $r_i = r(s_i, a_i)$, $r_j = r(s_j, a_j)$, $r_i' = r(s_i', a_i')$, $r_j' = r(s_j', a_j')$. For convenience, we did not indicate in the subscript of $\mathbb{E}$ the specific quantity being averaged.

If we further expand Eq.~\eqref{21}, all terms except for $c_n^k G_n^{c_n} - c_{n-1}^k G_{n-1}^{c_{n-1}}$ will contain $(c_{n-1} - c_n)$. 
According to Assumption~\ref{assumption_c}, terms that include $(c_{n-1} - c_n)$ tend to diminish towards $0$.
Additionally, since $c_n \in (0,1)$, as $k$ approaches $+\infty$, $c_n^k$ converges to $0$. 
Consequently, we can conclude that $|\mathcal{F}_{g}^{\pi}G_n^{c_n} - \mathcal{F}_{g}^{\pi}G_{n-1}^{c_{n-1}}| \rightarrow 0$.

For the seconde term:
\begin{equation}
\begin{aligned}
    &|\mathcal{F}_{g}^{\pi}G_n^{c_{n-1}} - \mathcal{F}_{g}^{\pi}G_{n-1}^{c_{n-1}}| \\
    =& | (1 - c_{n-1})|r_i - r_j| + c_{n-1} \mathbb{E}[U_n^{c_{n-1}}] - (1 - c_{n-1})|r_i - r_j| - c_{n-1} \mathbb{E}[U_{n-1}^{c_{n-1}}]| \\
    =& c_{n-1} | \mathbb{E}[ U_n^{c_{n-1}} - U_{n-1}^{c_{n-1}} ] | \\
    =& c_{n-1} |\mathbb{E}[ \mathbb{E}[G_n^{c_{n-1}} - G_{n-1}^{c_{n-1}}] ] | \\
    \leq& c_{n-1} \Vert G_n^{c_{n-1}} - G_{n-1}^{c_{n-1}} \Vert_{\infty},
\end{aligned}
\end{equation}
where $r_i = r(s_i, a_i)$, $r_j = r(s_j, a_j)$, $r_i' = r(s_i', a_i')$, $r_j' = r(s_j', a_j')$. For convenience, we did not indicate in the subscript of $\mathbb{E}$ the specific quantity being averaged.

Combining the two terms, we have:
\begin{equation}
    |G_{n+1} - G_n| = |\mathcal{F}_{g}^{\pi}G_n^{c_n} - \mathcal{F}_{g}^{\pi}G_{n-1}^{c_{n-1}}| \leq c_{n-1} \Vert G_n - G_{n-1} \Vert_{\infty}.
\end{equation}

Thus, we have proven that the operator $\mathcal{F}_{g}^{\pi}G$ in Eq.~\eqref{c-f} is a contraction mapping on $\mathbb{R}^{\mathcal{X}\times\mathcal{X}}$ with respect to $L^\infty$ norm under Assumption~\ref{assumption_c}.
    
\end{proof}
\end{proposition}

\section{Detailed Explanation of Toy Example}
\label{explanation of toy example}

Given a deterministic MDP $\mathcal{M} = (\mathcal{S}, \mathcal{A}, \mathcal{P}, \mathcal{R}, \gamma)$ and a policy $\pi$, where $\mathcal{S} = \{s_1, s_2, s_3\}$, $\mathcal{A} = \{a_0, a_1, a_2\}$, $\mathcal{R} = \{0, 1\}$, $\pi(a_0|s_1) = \pi(a_1|s_1) = \pi(a_0|s_2) = \pi(a_2|s_2) = \frac{1}{2}$, $\pi(a_0|s_3) = 1$, $r(s_1, a_0) = r(s_2, a_0) = 0$, $r(s_1, a_1) = r(s_2, a_2) = 1$, $r(s_3, a_0) = \frac{1}{2} $.

$\pi$-bisimulation operator~\cite{DBLP:conf/aaai/Castro20}:
\begin{equation}
    \mathcal{F}^{\pi}(d)(s_i, s_j)=|r_{s_i}^\pi-r_{s_j}^\pi|+\gamma \mathcal{W}(d)\left(P_{s_i}^\pi, P_{s_j}^\pi\right),
\end{equation} 
where $s_i,s_j\in \mathcal{S}$, $r_{s_i}^\pi=\sum_{a_i\in\mathcal{A}}\pi(a_i|s_i)r_{s_i}^{a_i}$ , $P_{s_i}^\pi=\sum_{a_i\in\mathcal{A}}\pi(a_i|s_i)P_{s_i}^{a_i}$, and $\mathcal{W}(d)$ is the Wasserstein distance with cost function $d$ between distributions.

Our new bisimulation operators:
\begin{equation}
\label{bisi-sa-u-appendix}
\begin{split}
    (\mathcal{F}_{u}^{\pi}U)(s_i, s_j) = \mathbb{E}_{\begin{subarray}{l} a_i \sim \pi(\cdot |s_i) \\ a_j \sim \pi(\cdot |s_j) \end{subarray}} \big[ G((s_i, a_i), (s_j, a_j)) \big],
\end{split}
\end{equation}
\begin{equation}
\label{bisi-sa-g-appendix}
\begin{split}
    (\mathcal{F}_{g}^{\pi}G)((s_i, a_i), (s_j, a_j)) = |r(s_i, a_i)-r(s_j, a_j)| + \gamma \mathbb{E}_{\begin{subarray}{l} s_i' \sim P(\cdot|s_i, a_i) \\ s_j' \sim P(\cdot|s_j, a_j) \end{subarray}} \big[ U(s_i', s_j') \big].
\end{split}
\end{equation}

\begin{figure*}[ht]
\vskip 0.2in
\begin{center}
     \includegraphics[width=0.45\textwidth]{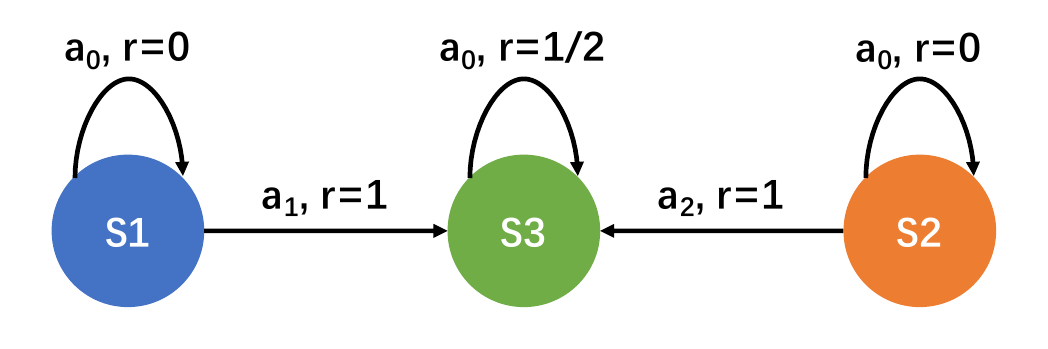}
     \caption{\textbf{Toy example.}}
\vskip -0.2in
\label{toy_example_appendix}
\end{center}
\end{figure*}

When using $\pi$-bisimulation operator to calculate the distance between each states:
\begin{equation}
\label{pi_s1_s2}
\begin{aligned}
    d(s1, s2) &= |[\pi(a_0|s_1)r(s_1,a_0) + \pi(a_1|s_1)r(s_1,a_1)] + [\pi(a_0|s_2)r(s_2,a_0) + \pi(a_2|s_2)r(s_2,a_2)]| \\
    &\quad + \gamma \{\pi(a_0|s_1)P(s_1'|s_1,a_0)\pi(a_0|s_2)P(s_2'|s_2,a_0)d(s_1',s_2') + \pi(a_0|s_1)P(s_1'|s_1,a_0)\pi(a_2|s_2)P(s_2'|s_2,a_2)d(s_1',s_2')\\
    &\quad+ \pi(a_1|s_1)P(s_1'|s_1,a_1)\pi(a_0|s_2)P(s_2'|s_2,a_0)d(s_1',s_2') + \pi(a_1|s_1)P(s_1'|s_1,a_1)\pi(a_2|s_2)P(s_2'|s_2,a_2)d(s_1',s_2')\} \\
    &= |[\pi(a_0|s_1)r(s_1,a_0) + \pi(a_1|s_1)r(s_1,a_1)] + [\pi(a_0|s_2)r(s_2,a_0) + \pi(a_2|s_2)r(s_2,a_2)]| \\
    &\quad + \gamma \{\pi(a_0|s_1)\pi(a_0|s_2)d(s_1,s_2) + \pi(a_0|s_1)\pi(a_2|s_2)d(s_1,s_3)\\
    &\quad+ \pi(a_1|s_1)\pi(a_0|s_2)d(s_3,s_2) + \pi(a_1|s_1)\pi(a_2|s_2)d(s_3,s_3)\} \\
    &= | (\frac{1}{2} \times 0 + \frac{1}{2} \times 1) - (\frac{1}{2} \times 0 + \frac{1}{2} \times 1) | \\
    &\quad + \gamma [ \frac{1}{2} \times \frac{1}{2} \times d(s_1,s_2) +  \frac{1}{2} \times \frac{1}{2} \times d(s_1,s_3) + \frac{1}{2} \times \frac{1}{2} \times d(s_3,s_2) + \frac{1}{2} \times \frac{1}{2} \times d(s_3,s_3)] \\
    &= 0 + \frac{\gamma}{4}  [ d(s_1,s_2) + d(s_1,s_3) + d(s_3,s_2) +d(s_3,s_3) ] \\
    &= \frac{\gamma}{4}  [ d(s_1,s_2) + d(s_1,s_3) + d(s_3,s_2) +d(s_3,s_3) ].
\end{aligned}
\end{equation}

Similarly, we can obtain the following equation:
\begin{equation}
\label{pi_s1_s3}
\begin{aligned}
    d(s_1, s_3) = \frac{\gamma}{2} [ d(s_1,s_3) + d(s_3,s_3) ],
\end{aligned}
\end{equation}
\begin{equation}
\label{pi_s3_s2}
\begin{aligned}
    d(s_3, s_2) = \frac{\gamma}{2} [ d(s_3,s_2) + d(s_3,s_3) ],
\end{aligned}
\end{equation}
\begin{equation}
\label{pi_s3_s3}
\begin{aligned}
    d(s_3, s_3) = \gamma d(s_3,s_3),
\end{aligned}
\end{equation}

Combine Eq.~\eqref{pi_s1_s2},~\eqref{pi_s1_s3},~\eqref{pi_s3_s2},~\eqref{pi_s3_s3}, and $\gamma \in (0,1)$, we have: $d(s_1,s_2)=d(s_1,s_3)=d(s_2,s_3)=d(s_3,s_3)=0$.

When using our new bisimulation operators to calculate the distance between each states:
\begin{equation}
\label{new-s1-s2}
\begin{aligned}
    U(s_1,s_2) &= \pi(a_0|s_1)\pi(a_0|s_2)G((s_1,a_0),(s_2,a_0)) + \pi(a_0|s_1)\pi(a_2|s_2)G((s_1,a_0),(s_2,a_2)) \\ 
    &\quad+ \pi(a_1|s_1)\pi(a_0|s_2)G((s_1,a_1),(s_2,a_0)) + \pi(a_1|s_1)\pi(a_2|s_2)G((s_1,a_1),(s_2,a_2)) \\
    &= \frac{1}{2} \times \frac{1}{2} G((s_1,a_0),(s_2,a_0)) + \frac{1}{2} \times \frac{1}{2} G((s_1,a_0),(s_2,a_2))\\
    &\quad + \frac{1}{2} \times \frac{1}{2} G((s_1,a_1),(s_2,a_0)) + \frac{1}{2} \times \frac{1}{2} G((s_1,a_1),(s_2,a_2)) \\
    &= \frac{1}{4} [ G((s_1,a_0),(s_2,a_0)) + G((s_1,a_0),(s_2,a_2)) + G((s_1,a_1),(s_2,a_0)) + G((s_1,a_1),(s_2,a_2)) ] \\
    &= \frac{1}{4} [ |r(s_1,a_0) - r(s_2,a_0)| + \gamma U(s_1,s_2) + |r(s_1,a_0) - r(s_2,a_2)| + \gamma U(s_1,s_3) \\
    &\quad + |r(s_1,a_1) - r(s_2,a_0)| + \gamma U(s_3,s_2) + |r(s_1,a_1) - r(s_2,a_2)| + \gamma U(s_3,s_3) ] \\
    &= \frac{1}{4} [|0 - 0| + \gamma U(s_1,s_2) + |0 - \frac{1}{2}| + \gamma U(s_1,s_3) + |\frac{1}{2} - 0| + \gamma U(s_3,s_2) + |\frac{1}{2} - \frac{1}{2}| + \gamma U(s_3,s_3) ] \\
    &= \frac{1}{4} [1 + \gamma (U(s_1,s_2) + U(s_1,s_3) + U(s_3, s_2) + U(s_3,s_3))]
\end{aligned}
\end{equation}

Similarly, we can obtain the following equation:
\begin{equation}
\label{new-s1-s3}
\begin{aligned}
    U(s_1,s_3) &= \frac{1}{2} [ G((s_1,a_0),(s_3,a_0)) + G((s_1,a_1),(s_3,a_0)) ]\\
    &= \frac{1}{2}[ |r(s_1,a_0) - r(s_3,a_0)| + \gamma U(s_1,s_3) + |r(s_1,a_1) - r(s_3,a_0)| + \gamma U(s_3,s_3) ]\\
    &= \frac{1}{2}[ |0 - \frac{1}{2}| + \gamma U(s_1,s_3) + |\frac{1}{2} - \frac{1}{2}| + \gamma U(s_3,s_3) ]\\
    &= \frac{1}{2} [ \frac{1}{2} + \gamma (U(s_1,s_3) + U(s_3,s_3)) ],
\end{aligned} 
\end{equation}
\begin{equation}
\label{new-s3-s2}
\begin{aligned}
    U(s_3,s_2) &= \frac{1}{2} [ G((s_3,a_0),(s_2,a_0)) + G((s_3,a_0),(s_2,a_2)) ]\\
    &= \frac{1}{2}[ |r(s_3,a_0) - r(s_2,a_0)| + \gamma U(s_3,s_2) + |r(s_3,a_0) - r(s_2,a_2)| + \gamma U(s_3,s_3) ]\\
    &= \frac{1}{2}[ |\frac{1}{2} - 0| + \gamma U(s_3,s_2) + |\frac{1}{2} - \frac{1}{2}| + \gamma U(s_3,s_3) ]\\
    &= \frac{1}{2} [ \frac{1}{2} + \gamma (U(s_3,s_2) + U(s_3,s_3)) ],
\end{aligned} 
\end{equation}
\begin{equation}
\label{new-s3-s3}
\begin{aligned}
    U(s_3,s_3) &= G((s_3,a_0),(s_3,a_0)) \\
    &= |r(s_3,s_3) - r(s_3,s_3)| + \gamma U(s_3,s_3) \\
    &= |\frac{1}{2} - \frac{1}{2}| + \gamma U(s_3,s_3) \\
    &= \gamma U(s_3,s_3).
\end{aligned} 
\end{equation}

Combine Eq.~\eqref{new-s1-s2},~\eqref{new-s1-s3},~\eqref{new-s3-s2},~\eqref{new-s3-s3}, and $\gamma \in (0,1)$, we have: $U(s_1,s_2)=\frac{2}{(2-\gamma)(4-\gamma)}$, $U(s_1,s_3)=\frac{1}{4-2\gamma}$, $U(s_3,s_2)=\frac{1}{4-2\gamma}$, $U(s_3,s_3)=0$.

\section{Notation}

Table \ref{table-notation} summarizes our notation.

\begin{table}[hbp]
\vskip 0.15in
\centering
\caption{Table of Notation. }
\begin{center}
\begin{small}
\begin{tabular}{cccc}
\toprule
Notation                    & Meaning                                  & Notation                    & Meaning                            \\ \midrule
$\mathcal{M}$ & MDP  & $\mathcal{S}$ & state space \\ \midrule
$\mathcal{A}$ & action space  & $\mathcal{X}$ & state-action space \\ \midrule
$\mathcal{P}$ & transition function  & $\mathcal{R}$ & reward function \\ \midrule
$\gamma$ & discount factor  & $c$ & weight coefficient \\ \midrule
$\pi$ & policy of the agent  & $\mathcal{D}$ & replay buffer \\ \midrule
$V^{\pi}(s)$ & state value function given policy $\pi$  & $Q^{\pi}(s,a)$ & state-action value function given policy $\pi$ \\ \midrule
$\mathcal{R}$ & reward function  & $\gamma$ & discount factor \\ \midrule
$U(\cdot,\cdot)$ & a specific distance function between states  & $G(\cdot,\cdot)$ & a specific distance function between state-action pairs \\ \midrule
$\mathcal{F}^{\pi}_{u}$ & on-policy bisimulation operator for $U$  & $\mathcal{F}^{\pi}_{g}$ & on-policy bisimulation operator for $G$ \\ \midrule
$\phi$ & state encoder & $\psi$ & state-action encoder \\ 
\bottomrule
\end{tabular}
\end{small}
\end{center}
\vskip -0.1in
\label{table-notation}
\end{table}

\newpage
\section{Algorithms}
\setcounter{algorithm}{0}
\begin{algorithm}[htbp]
\caption{MICo+SA+C Pseudocode, PyTorch-like}
\label{algo-mico-sa-c}
\definecolor{codeblue}{rgb}{0.25,0.5,0.5}
\definecolor{codekw}{rgb}{0.85, 0.18, 0.50}
\lstset{
  backgroundcolor=\color{white},
  basicstyle=\fontsize{7.5pt}{7.5pt}\ttfamily\selectfont,
  columns=fullflexible,
  breaklines=true,
  captionpos=b,
  commentstyle=\fontsize{7.5pt}{7.5pt}\color{codeblue},
  keywordstyle=\fontsize{7.5pt}{7.5pt}\color{codekw},
}

\begin{lstlisting}[language=python]
# encoder: cnn + mlp, state encoder network, the output is L2-normalized
# project_sa: mlp, state-action encoder network, the output is L2-normalized
# raw_c = torch.tensor([0.0, 0.0])

def cosine_dis(feature_a, feature_b):
    feature_a_norm = nn.functional.normalize(feature_a, dim=1)
    feature_b_norm = nn.functional.normalize(feature_b, dim=1)
    temp = torch.matmul(feature_a_norm, feature_b_norm.T)
    dis = torch.diag(temp)
    
    return dis

def diff(feature_a, feature_b, feature_cosine_dis, beta=0.1):
    feature_cosine_dis = torch.clip(feature_cosine_dis, min=0.0001, max=0.9999)
    feature_atan2_dis = torch.atan2(torch.sqrt(torch.ones_like(feature_cosine_dis)
                    - torch.pow(feature_cosine_dis, 2)), feature_cosine_dis)
    diff = 0.5 * (torch.norm(feature_a, dim=1) + torch.norm(feature_b, dim=1)) + beta * feature_atan2_dis
    
    return diff
    
def compute_loss(encoder, target_encoder, replay_buffer, batch_size, discount, slope):
    # sample a batch of tuples from replay buffer
    observation, action, reward, discount, next_observation = replay_buffer.sample(batch_size) 
    
    latent_state = encoder(observation)
    latent_state_action = project_sa(latent_state, action)
    with torch.no_grad():
        latent_next_state = encoder(next_observation)
        
    r_diff = torch.abs(reward.T - reward)
    next_diff = compute_distance(latent_next_state, latent_next_state)
    z_diff = compute_distance(latent_state_action, latent_state_action)
    c = torch.nn.functional(self.raw_c, dim=0)
    c = F.softmax(self.raw_c, dim=0)
    bisimilarity = c[0] * r_diff + c[1] * next_diff
    
    encoder_loss = torch.nn.HuberLoss()(z_diff, bisimilarity.detach())
    c_loss = torch.nn.HuberLoss()(z_diff.detach(), bisimilarity)
    
    return encoder_loss, c_loss
\end{lstlisting}
\end{algorithm}

\begin{algorithm}[h!]
\caption{SimSR+SA+C Pseudocode, PyTorch-like}
\label{algo-simsr-sa-c}
\definecolor{codeblue}{rgb}{0.25,0.5,0.5}
\definecolor{codekw}{rgb}{0.85, 0.18, 0.50}
\lstset{
  backgroundcolor=\color{white},
  basicstyle=\fontsize{7.5pt}{7.5pt}\ttfamily\selectfont,
  columns=fullflexible,
  breaklines=true,
  captionpos=b,
  commentstyle=\fontsize{7.5pt}{7.5pt}\color{codeblue},
  keywordstyle=\fontsize{7.5pt}{7.5pt}\color{codekw},
}

\begin{lstlisting}[language=python]
# encoder: cnn + mlp, state encoder network, the output is L2-normalized
# project_sa: mlp, state-action encoder network, the output is L2-normalized
# raw_c = torch.tensor([0.0, 0.0])

def compute_distance(features_a, features_b):
    similarity_matrix = torch.matmul(features_a, features_b.T)
    dis = 1-similarity_matrix
    
    return dis
    
def compute_loss(encoder, target_encoder, replay_buffer, batch_size, discount, slope):
    # sample a batch of tuples from replay buffer
    observation, action, reward, discount, next_observation = replay_buffer.sample(batch_size) 
    
    latent_state = encoder(observation)
    latent_state_action = project_sa(latent_state, action)
    with torch.no_grad():
        latent_next_state = encoder(next_observation)
        
    r_diff = torch.abs(reward.T - reward)
    next_diff = compute_distance(latent_next_state, latent_next_state)
    z_diff = compute_distance(latent_state_action, latent_state_action)
    c = torch.nn.functional(self.raw_c, dim=0)
    c = F.softmax(self.raw_c, dim=0)
    bisimilarity = c[0] * r_diff + c[1] * next_diff
    
    encoder_loss = torch.nn.HuberLoss()(z_diff, bisimilarity.detach())
    c_loss = torch.nn.HuberLoss()(z_diff.detach(), bisimilarity)
    
    return encoder_loss, c_loss
\end{lstlisting}
\end{algorithm}

\section{Environment Details}
\textbf{DeepMind Control Suite.}
The DeepMind Control Suite (DMControl)~\cite{tassa2018deepmind} is a widely adopted benchmark for continuous control tasks, providing a diverse set of environments to evaluate the performance of reinforcement learning algorithms. We leverage this benchmark to validate the generalization capability and robustness of our proposed method across a range of challenging control tasks. Specifically, we select a set of nine relatively difficult tasks that vary in complexity, action space, and reward density, ensuring a comprehensive evaluation of our approach.

\textbf{Meta-World.}
To evaluate our method's ability to accelerate policy optimization when jointly trained with the policy, we conduct experiments on robotic arm manipulation tasks in the Meta-World environment~\cite{yu2020meta}, a benchmark for multi-task and meta-reinforcement learning. Each task includes 50 randomized configurations, varying in robot poses, object locations, and target positions to mimic real-world variability and test generalization. We select three tasks with increasing complexity (simple, medium, and hard) to analyze the algorithm's performance across different task difficulties.

\section{Experimental Details}
\subsection{Implementation Details}
In our experiments on DMC, we follow the publicly available implementation of~\citet{zheng2023taco}. For the Meta-World experiments, we adhere to the publicly available implementation of DrQ-v2 provided by~\citet{xu2023drm}. To further evaluate the performance of MICo~\cite{DBLP:conf/nips/CastroKPR21} and SimSR~\cite{zang2022simsr} across different types of tasks, while ensuring fairness and comparability of experimental results, we standardize all methods within the same code framework and strictly adopted consistent hyperparameter settings. All networks are optimized using the Adam optimizer~\cite{kingma2014adam}.
\subsection{Hyperparameter}
\begin{table}[h!]
  \caption{A default set of hyperparameters used in our experiments.
  }
  \centering
  \begin{tabular}{lll}
  \toprule
  \textbf{Hyperparameter} & DMControl & Meta-World\\
  \midrule
  \multicolumn{2}{l}{\textbf{General}} \\
  \midrule
  Batch size & 1024 & 1024\\
  Replay buffer capacity & 1000000 & 1000000\\ 
  Action repeat & 2 & 2\\ 
  Seed frames & 4000 & 4000\\ 
  Exploration steps & 2000 & 2000\\ 
  n-step returns & 3 & 3\\ 
  Discount & 0.99 & 0.99 \\
  Actor update frequency & $2$ & $2$\\
  Critic Q-function soft-update rate  & $0.01$ & $0.01$\\
  State representation dimension & 50 & 50 \\
  Exploration stddev. clip & $0.3$ & $0.3$\\
  Exploration stddev. schedule & linear(1.0, 0.1, 2000000) & linear(1.0, 0.1, 3000000)\\
  Init temperature & $0.1$ & $0.1$\\
  Learning rate & $1\times10^{-4}$ & $1\times10^{-4}$\\
  Temperature learning rate & $1\times10^{-4}$ & $1\times10^{-4}$\\
  \midrule
  \multicolumn{2}{l}{\textbf{Bisimulation}} \\
  \midrule
  State-Action representation dimension & 50 & 50 \\
  Initial values of $c1$ and $c2$ & 0.5 & 0.5 \\
  Representation learning rate & $1\times10^{-4}$ & $1\times10^{-4}$\\
  Learning rate of $c$ & $1\times10^{-4}$ & $1\times10^{-4}$\\
  \bottomrule
  \end{tabular}
  \label{sac:hparams}
  \end{table}
\end{document}